\documentclass[10pt,twocolumn,letterpaper]{article}

\usepackage[pagenumbers]{cvpr} 

\definecolor{cvprblue}{rgb}{0.21,0.49,0.74}
\usepackage[pagebackref,breaklinks,colorlinks,allcolors=cvprblue]{hyperref}

\newtheorem{definition}{Definition}

\newcommand{\clustering}[1]{\mathcal{#1}}
\newcommand{\R}{\mathbb{R}}

\title{Beyond Scalars: Concept-Based Alignment Analysis in Vision Transformers}

\author{Johanna Vielhaben$^1$ \and
        Dilyara Bareeva$^{1}$ \and
        Jim Berend$^{1}$ \and
        Wojciech Samek$^{1,2,3,\dagger}$ \and
        Nils Strodthoff$^{4,\dagger}$ 
        \and 
\footnotesize$^1$ Fraunhofer Heinrich-Hertz-Institute Berlin\;
\footnotesize$^2$ Technische Universität Berlin\;
\footnotesize$^3$ BIFOLD – Berlin Institute for the Foundations of Learning and Data \\
\footnotesize$^4$ Carl von Ossietzky Universität Oldenburg \;
\footnotesize$^\dagger$ corresponding authors: \texttt{wojciech.samek@hhi.fraunhofer.de, nils.strodthoff@uol.de}
}

\begin{document}
\maketitle

\begin{abstract}
Vision transformers (ViTs)~\cite{dosovitskiy2021an} can be trained using various learning paradigms, from fully supervised to self-supervised. Diverse training protocols often result in significantly different feature spaces, which are usually compared through alignment analysis. However, current alignment measures quantify this relationship in terms of a single scalar value, obscuring the distinctions between common and unique features in pairs of representations that share the same scalar alignment.
We address this limitation by combining alignment analysis with concept discovery, which enables a breakdown of alignment into single concepts encoded in feature space. 
This fine-grained comparison reveals both universal and unique concepts across different representations, as well as the internal structure of concepts within each of them.
Our methodological contributions address two key prerequisites for concept-based alignment: 1) For a description of the representation in terms of concepts that faithfully capture the geometry of the feature space, we define concepts as the most general structure they can possibly form - arbitrary manifolds, allowing hidden features to be described by their proximity to these manifolds. 2) To measure distances between concept proximity scores of two representations, we use a generalized Rand index 
and partition it for alignment between pairs of concepts.
We confirm the superiority of our novel concept definition 
for alignment analysis over existing linear baselines in a sanity check. The concept-based alignment analysis of representations from four different ViTs reveals that increased supervision correlates with a reduction in the semantic structure of learned representations. 
\end{abstract}

\section{Introduction}
Vision Transformers are gaining increased popularity as backbones for various computer vision tasks.
There is a large zoo of pre-trained models trained with various learning paradigms and a range of supervision strengths. To guide practitioners, previous work has evaluated performance on various common downstream tasks~\cite{Goldblum2023BattleOT}. 
\begin{figure}[ht]
    \centering
    \includegraphics[width=\linewidth]{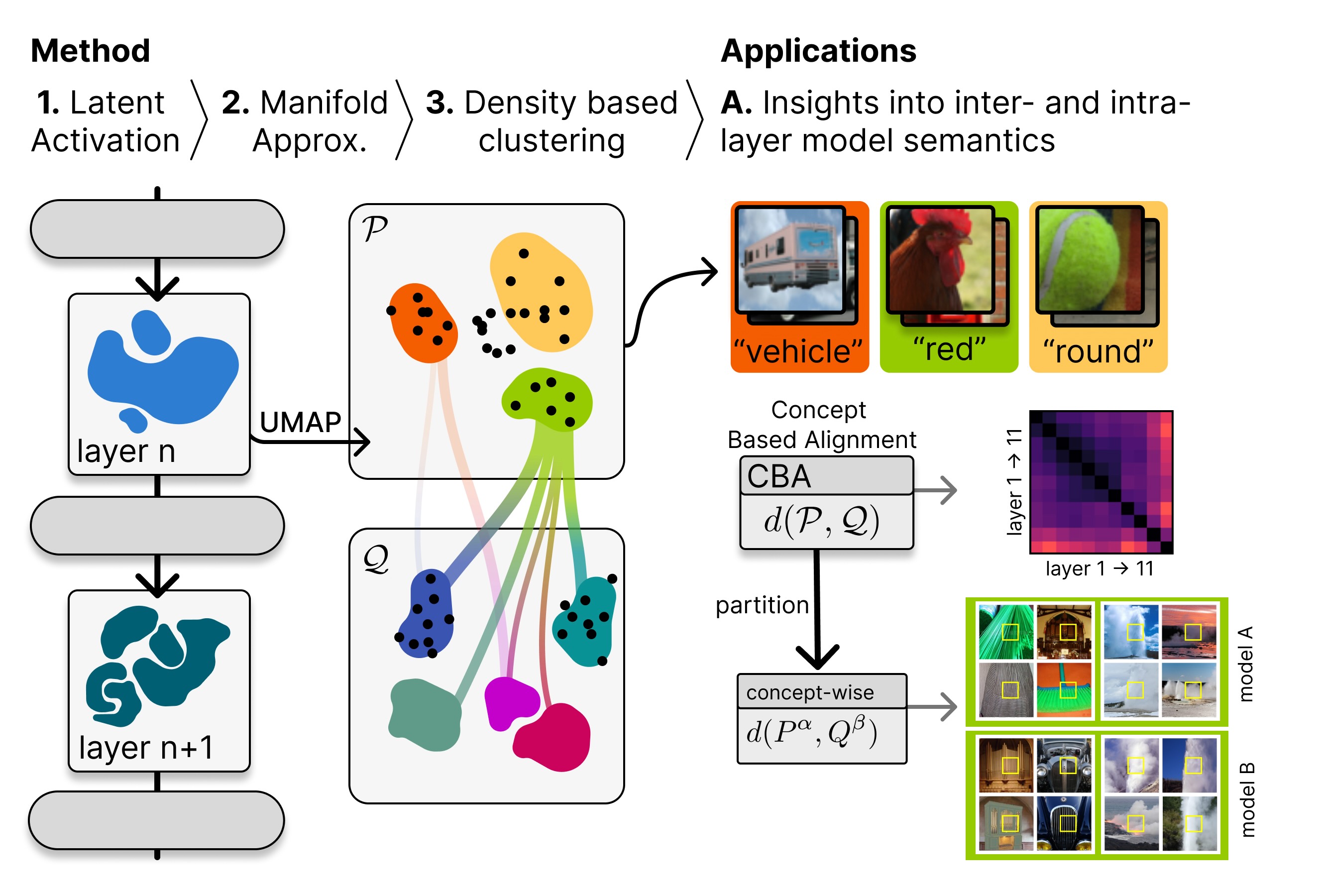}
    \caption{We combine concept discovery with alignment analysis for fine-grained insights into structures within and differences between latent activations. To this end, we investigate latent activations formed by intermediate layers, which according to the manifold hypothesis can be organized in terms of low-dimensional manifolds. We recover manifolds using density-based clustering applied to UMAP embeddings of the latent representations. The discovered structures in latent space do not only allow to characterize a single layer, but also the formation of structures between layers.}
    \label{fig:teaser}
\end{figure}
A complimentary view of comparisons within and between models beyond quantitative accuracy is achieved by analyzing patterns in hidden activations and measuring representational alignment between them~\cite{Raghu2021DoVT, Walmer2022TeachingMI}.

When choosing a model for a downstream task,
we want to understand how the model solves its pre-training task. Where does the model representation change the most and how? Which concepts,  i.e. dominant structures in representation space, are encoded in lower layers vs. upper layers?  Where does the model representation change the most and how? How structured are the representations? Does the model encode semantically similar concepts in spatial proximity to each other? How is the representation of model A different from that of model B across layers?
Answering these questions can aid the selection of pre-trained models and the design of fine-tuning strategies through detailed insights into robustness and generalization capabilities.
 Previous work on alignment, however, only provides a single scalar value to measure alignment between representations at two different layers~\cite{Sucholutsky2023GettingAO}, leaving the questions above largely unanswered.
In this paper, we propose a more fine-grained alignment analysis based on concepts that structure the latent representation. To this end, we represent the original activations by concept membership scores that quantify proximity to the discovered concepts. Then, we measure alignment between concept proximity scores of representations and can therefore partition it into the concepts. This gives insights into universal and specific concepts between representations of different layers or models, as well as how a single representation is structured.

To achieve concept-based alignment we need solutions for 1) \textcolor{violet}{concept discovery}, and  2) measuring the \textcolor{olive}{alignment} between concept proximity scores.

Previous work on \textcolor{violet}{concept discovery} ranges from merely identifying neurons or other pre-existing units as concepts~\cite{BauNetDissect} to linear directions in feature space~\cite{fel2023holistic}. The most general definition so far relies on concepts as multi-dimensional linear subspaces~\cite{Vielhaben2023}. The common strong assumption among these is the linearity of concept structures, which is challenging to verify and controversial~\cite{bereska2024mechanistic, csordas2024RNNnonlinear}. For concepts that faithfully represent the underlying geometry of the representation, we avoid the linearity assumption and consider concepts as the most general structure they can form, namely as nonlinear manifolds.
So far, \textcolor{olive}{alignment} between representations has been measured as the similarity of similarities, e.g.\ through linear or kernel-based Centred Kernel Alignment (CKA)~\cite{Kornblith2019SimilarityON}, which results in a single scalar value. Our fine-grained concept-based alignment measure requires a distance measure between concept proximity scores. Here, we choose a generalized Rand index between soft clusterings with pseudo metric properties~\cite{hullermeier2012fuzzy} that we partition into pairwise concept distances.

To summarize, our key idea is the following:
\begin{itemize}
    \item We combine \textcolor{violet}{concept discovery} with \textcolor{olive}{alignment} analysis to provide insights into which concepts are universal or specific between two representations, and how structured a single representation is.
\end{itemize}
We make the following methodological contributions to realize concept-based alignment:
\begin{itemize}
    \item We propose a novel concept definition of concepts as nonlinear manifolds to faithfully capture the geometry of the feature space with concept proximity scores.
    \item We leverage a generalized Rand index with pseudo-metric properties to measure the alignment between concept proximity scores of two representations and partition it for fine-grained concept alignment.
\end{itemize}
We complement concept-based alignment analysis of ViTs trained under varying degrees of supervision from fully supervised to self-supervised with additional characteristics of concepts such as their intrinsic dimensionality.
We find that representations of ViTs exhibit markedly different structures; specifically, increased supervision correlates with reduced structure in the learned representations. This insight is crucial for understanding the model's reasoning processes and sheds light on the performance differences observed in quantitative analyses, such as those presented in the recent battle-of-the-backbones study~\cite{Goldblum2023BattleOT}. Code to reproduce our experiments is publicly available at \url{https://github.com/jvielhaben/NLMCD-ALIGN}.

\section{Concept Discovery for Representational Alignment}
This section is partitioned into three parts: First, we introduce our novel concept definition based on the manifold hypothesis. Then, we describe our methodology for discovering these concepts in latent activations, shown in Fig.~\ref{fig:teaser}. Finally, we describe how our concept-based description of hidden representations can be used to measure alignment between representations, identify commonalities and uniqueness between models, and investigate information flow within one model.

\subsection{Concept definition}
\paragraph{Motivation}
According to the manifold hypothesis, which is widely accepted in machine learning, many datasets, including image data that nominally lie in high dimensional space, can be described in terms of a few underlying latent factors and are thus concentrated on a (potentially disconnected) low-dimensional manifold embedded in high-dimensional space~\cite{Goodfellow2016Deep}.~\cite{Naitzat2020} shows how a neural network trained on a toy classification problem solves the task by transforming the topology of the input data, and layerwise reducing the Betti numbers of the class-wise components. We hypothesize that state-of-the-art vision models behave similarly and try to recover the connected components in the hidden representations, which we call \emph{concepts}.
\vspace{-0.4cm}
\paragraph{Definition}
We analyze the hidden representation at an intermediate feature layer of a neural network. To this end, we split the model $f$ into two parts, $f = g_l \circ h_l$, where $h_l$ is the mapping to a hidden feature layer $l$. Our definition then relies on hidden representations $h_l(x_i) \in \R^{N' \times F }$ of input samples $x_i$ from a set $S$. $N'$ is the number of spatially separable elements in the representation, i.e.\ the number of tokens in a transformer model or the number of superpixels in a convolutional feature map. We spatially decompose the feature maps $h(x_i)$ into a set of $N=N'\cdot |S|$ feature vectors $\phi \in \R^F$.
Previously, concepts have been mostly defined as linear structures~\cite{fel2023holistic, Zhang2021Invertible}. The most general linear structure would be affine subspaces, which would already represent an extension compared to the recently considered definition as linear subspaces~\cite{Vielhaben2023}.
In this work, we generalize this idea even one step further and define concepts as manifolds in the $F$-dimensional feature space. 
\begin{definition}
    We define a concept $C^{\alpha}$, as a manifold in the $F$-dimensional feature space, represented by a point cloud $\{\phi^{\alpha}_j\}$ consisting of the feature vectors $\phi_j$ that lie on the concept manifold with index $\alpha$.
\end{definition}
\vspace{-0.4cm}
\paragraph{Benefits of concept manifold definition}
 In the following, we want to compute concept proximity scores by which we measure alignment. Incorrect assumptions about the structure of the concept manifold, e.g., assuming it has no curvature (affine subspaces) or it is spherical and the distance to the manifold can be estimated by the distance to the centroid, directly lead to distorted concept proximity scores and hence to distorted alignment.
Later, in a sanity check our definition performs best for measuring representational alignment. 

\subsection{Concept discovery} \label{sec:methods_concept_discovery}
\paragraph{Clustering}
Having established our definition of concepts as manifolds in feature space, we now turn to the challenge of discovering these concepts through clustering.
As stated above, we assume that feature vectors $\{\phi_i\}$ from a hidden representation are sampled from a set of low-dimensional concept manifolds $\{C^{\alpha}\}$.
Recovering these concept manifolds in high-dimensional space ($F=768$ in our experiments) is a challenging clustering problem. Therefore, we revert to density-based clustering on a low-dimensional embedding of the data~\cite{Goh2008,Herrmann2023}. 
For this embedding, we utilize UMAP (Uniform Manifold Approximation and Projection)~\cite{McInnes2018UMAPUM}, a dimensionality reduction technique that preserves local and some global structure. 
Given that we have no a priori knowledge about the number of clusters, we employ HDBSCAN (Hierarchical Density-Based Spatial Clustering of Applications with Noise), which can handle clusters of varying densities~\cite{Campello2013DensityBasedCB}. HDBSCAN builds a hierarchy of clusters based on density, represented by a condensed tree, and allows for robust handling of noise, making it suitable for the possibly intricate structure of feature representation spaces. While UMAP does not fully preserve density, its ability to maintain the overall structure of the data makes it a valuable preprocessing step before applying HDBSCAN. We use the HDBSCAN implementation from~\cite{McInnes2017}.
\paragraph{Concept proximity scores}
We leverage soft clustering with HDBSCAN based on the condensed tree which is roughly a density function over the data points to compute fuzzy cluster membership as described in~\cite{McInnes2017}, which we formalize in the appendix for the reader's convenience. It is based on the distance to concept anchor points a cluster and an outlier score, both derived from the condensed tree. We now have a fuzzy clustering $\clustering{P}_{\{\phi\}}= \{\clustering{P}(\phi_0), \dots, \clustering{P}(\phi_N)\}$, where $\clustering{P} \in [0,1]^n$ holds the concept proximity scores of each concept $C^{\alpha}$. We interpret the concept proximity scores $P^{\alpha}(\phi)$ as the probability that a feature vector $\phi$ belongs to a concept $P^{\alpha}$ in clustering $\clustering{P}$.
This approach contrasts with previous concept assignment paradigms~\cite{fel2023holistic, Vielhaben2023}, which often rely on hard clustering, where each feature vector is assigned to a single concept, or linear methods that project onto specific concept directions, limiting the representation to a more rigid framework. In contrast, our soft clustering method allows for nuanced membership scores that reflect the degree of belonging to multiple concepts.
In the following, we refer to our concept discovery method as \emph{NLMCD} (non-linear multi-dimensional concept discovery).

\subsection{Concept-based Representational Alignment} \label{sec:methods_alignment}
We now address the question of measuring representational alignment based on the concept proximity scores derived from fuzzy clustering.
\vspace{-0.4cm}
\paragraph{Pseudo-metric between fuzzy clusterings} The concepts are at this point characterized by a probabilistic clustering $\clustering{P}_{\{\phi\}}= \{\clustering{P}(\phi_0), \dots, \clustering{P}(\phi_N)\}$, where $\clustering{P}(\phi_i)=\{ P^{1}(\phi_i), \dots , P^{n}(\phi_i) \}$. We want to measure the similarity between two probabilistic clusterings $\clustering{P}, \clustering{Q}$ from two different representations to evaluate how aligned their concepts are. For this purpose, we leverage an extension of the pair-based Rand index generalized to fuzzy clusterings proposed in~\cite{hullermeier2012fuzzy}. The original Rand index counts the number of concordant pairs (either two points are paired or not paired both clusterings) and disconcordant pairs (two points are paired in one clustering but not in the other). The distance between probabilistic clustering $\clustering{P}, \clustering{Q}$ is based on a generalized degree of concordance that is based on the \textit{distance between two membership vectors} $d_{ms}(\clustering{P}(\phi_i),\clustering{P}(\phi_j))$:
\begin{align}\label{eq:CBA_dist}
    d_{cross}(\clustering{P}, \clustering{Q}) &=  \frac{2}{N(N-1)} \sum_{i,j} \lvert d_{ms}(\clustering{P}(\phi_i),\clustering{P}(\phi_j)) \\ &- d_{ms}(\clustering{Q}(\phi_i),\clustering{Q}(\phi_j)) \rvert   
\end{align} 
 A commonly used choice for the distance $d_{ms}$ is  $d_{ms} (P(\phi_i),P(\phi_j))=1-||P(\phi_i)-P(\phi_j)||_1$~\cite{DeWolfe2023RandomMF}.
Finally, we refer to the similarity between two clusterings, derived from the uncovered concepts, as \textit{Concept-Based Alignment} (CBA):
 \begin{equation} \label{eq:CBA}
     \text{CBA} = 1 - d_{cross}(\clustering{P}, \clustering{Q})
 \end{equation}
 We choose this measure because $d_{cross}(\clustering{P}, \clustering{Q})$ is a pseudo-metric satisfying desirable properties\footnote{1) Identity: \(d(x, x) = 0\) for all \(x\), 2) Symmetry: \(d(x, y) = d(y, x)\) for all \(x, y\), 3) Triangle Inequality: \(d(x, z) \leq d(x, y) + d(y, z)\) for all \(x, y, z\).} that ease interpretation Also, when $\clustering{P}, \clustering{Q}$ are crisp partitions, CBA reduces to the original Rand index.

\paragraph{Similarity index between clusters} In contrast to conventional measures for representational alignment that yield a single scalar value, our approach provides a more nuanced measure of representational alignment by assessing similarities and differences between clusters.  To measure distance between two clusters $P^{\alpha}$, $Q^{\beta}$
from two clusterings $\clustering{P}$, $\clustering{Q}$, we decompose the distance in Eq.~\ref{eq:CBA_dist} into the contribution of single concepts $P^{\alpha}$, $Q^{\beta}$ and measure the \textit{pairwise similarity between the membership scores} of each feature
\begin{align} \label{eq:CBA_dist_c}
    d_{cross}(P^{\alpha}{,}Q^{\beta} ) =  \tfrac{2}{N^2{-}N} &\sum_{i,j} \lvert \lvert P^{\alpha}(\phi_i)-P^{\alpha}(\phi_j)\rvert \\ \nonumber &- \lvert Q^{\beta}(\phi_i)-Q^{\beta}(\phi_j)\rvert \rvert 
\end{align}
Due to the absolute value in Eq.~\ref{eq:CBA_dist}, summing over all pairs $\alpha,\beta$ does not yield the total $d_{cross}(\clustering{P}, \clustering{Q})$, but by the triangle inequality $\sum_{\alpha,\beta}d_{cross}(P^{\alpha},Q^{\beta} )  \geq d_{cross}(\clustering{P}, \clustering{Q}) $ the sum is an upper bound for the overall distance between two clusterings.

\section{Related work}
\paragraph{Concept discovery}
Most existing methods model concepts as linear directions~\cite{ghorbani2019towards,Zhang2021Invertible,fel2023craft, fel2023holistic}. Generalizing this definition,~\cite{Vielhaben2023} suggest that concepts can be represented more faithfully as multidimensional linear subspaces, which they discover through sparse subspace clustering. While above methods operate unsupervised without concept labels,~\cite{crabbe2023concept} employ kernel classifier for supervised, nonlinear concept discovery, showing improvement over linear concepts. In the field of \emph{mechanistic interpretability}, many studies aim to enumerate all \emph{features} encoded in the representations of neural networks~\cite{bricken2023monosemanticity}. This line of work focuses mainly on language models, often identifying linear features using sparse autoencoders~\cite{marks2024sparsefeaturecircuitsdiscovering,huben2024sparse,gao2024scalingevaluatingsparseautoencoders}. However,~\cite{Engels2024NotAL} find evidence for the existence of multi-dimensional non-linear features. Unlike these approaches, our main goal in concept discovery is representation summarization for alignment measurement, rather than interpretability or feature enumeration. For this reason, we employ the most general, non-linear concept definition.

\paragraph{Alignment}
Representational alignment measures are categorized, with a particular emphasis on Centered Kernel Alignment (CKA) in \cite{Kornblith2019SimilarityON}. CKA evaluates the similarity of similarities, either linearly or under a non-linear kernel. Similarly, \cite{Doimo2020} measure alignment through the similarities of binary k-nearest neighbor adjacency matrices, which resembles CKA with a narrow Gaussian kernel. 
Our method relates to CKA in that it condenses these similarities into clusters and subsequently measures the similarity between these clusterings. 

\paragraph{Comparison of Vision Models}
On the one hand, alignment measures such as CKA have been used to compare the representations of various architectures, including ViTs and ResNets trained on different tasks, together with the analysis of patterns in attention maps \cite{Walmer2022TeachingMI, Raghu2021DoVT}. Further, the analysis of attention patterns reveals differences between self-supervised ViTs \cite{Park2023WhatDS}.
On the other hand, downstream performance is analyzed to guide the selection of pre-trained models for transfer learning. Through this, \cite{Kornblith2018DoBI} shows that models pre-trainined on ImageNet generalize well but when used as feature extractors in transfer learning, i.e. when weights are completely frozen, perform badly in some settings, suggesting that the features of the last layers do not generalize well.
An extensive evaluation of the downstream performance of a large selection of vision models on classification, detection, image retrieval, and generalization is available in \cite{Goldblum2023BattleOT}.

\section{Results}
We evaluate concept discovery in Sec.~\ref{sec:results_concept_discovery}, check the superiority of our new concept definition over linear baselines for concept alignment analysis in Sec.~\ref{sec:sanity_check}, and perform a concept-alignment analysis between four ViTs in Sec.~\ref{sec:results_alignment}.

\subsection{Concept discovery} \label{sec:results_concept_discovery}
\begin{figure}[t]
    \centering
    \includegraphics[width=1.0\linewidth]{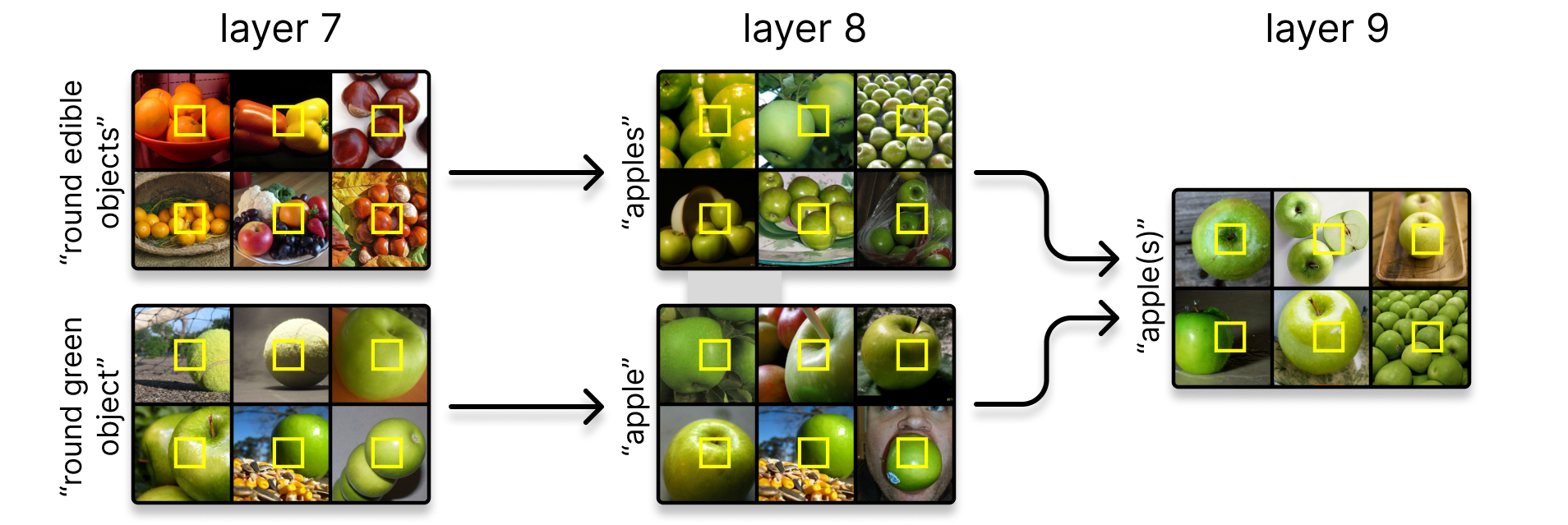}
    \caption{Concept formation graph for the concept ``apple(s)'' in layer 9 of the FS model. Each concept is represented by six randomly sampled images containing a token assigned to that concept  (highlighted in a yellow frame). 
    }
    \label{fig:concept_formation}
\end{figure}

First, we outline the concept discovery procedure as described in Sec.~\ref{sec:methods_concept_discovery} and evaluate the quality of the UMAP embeddings used for HDBSCAN clustering and the clustering itself.
\begin{table}[]
    \centering
    \begin{tabular}{@{} p{0.7cm} p{1.95cm} p{4.8cm} @{}}
    \toprule
         Model & Training Data & Training Task  \\
         \midrule
         FS \cite{steiner2022how} & ImageNet-1k \cite{russakovsky2015the} & Fully supervised learning with labeled data for classification task. \\
         CLIP \cite{radford2021learning} & WebImageText  \cite{radford2021learning} & Contrastive learning between images and text.\\
         DINO \cite{caron2021emerging} & ImageNet-1k &
         Knowledge distillation enforcing consistency between augmented views of the same image. \\
         MAE \cite{he2022masked} & ImageNet-1k & Masked autoencoders to reconstruct missing pixels of input data. \\
         \bottomrule
    \end{tabular}
        \caption{Pre-trained models we study with concept-based alignment, which range from fully supervised to text-image contrastive to self-supervised. Sources for the model weights are provided in the appendix.}
    \label{tab:models}
\end{table}
For concept discovery and later analysis of representational alignment, we use a random subset of 25 \% of the ImageNet train set, stratified samples across all 1000 classes.
We study four different ViTs~\cite{dosovitskiy2021an} with the same architecture (base, patch size 16, input size 224) but different training objectives and training datasets described in \cref{tab:models}.
We perform concept discovery separately for the sequence (SEQ) and the CLS token. We extract activations at the last MLP layer of each of the twelve transformer blocks. For the sequence tokens, we average-pool $4 \times 4$ token and select one of the pooled tokens from the sequence with more weight on the center of the image. For SEQ tokens, we discard the last block as for the considered models only the CLS token in the final layer enters the loss.
We evaluate how well the embedding on which we perform the clustering preserves the distances by measuring the mean squared error between the distance matrices in the original representation and its embedding (RMSE). 
To evaluate the clustering, we compute a density-based validity index (DBCV)~\cite{Moulavi2014DensityBasedCV}, which measures intra- vs inter-cluster density. Further, we report the rate of points classified as noise by HDBSCAN. To treat the noise rate and validity index separately, we do not weight the average for the DBCV across clusters by the cluster size as proposed in~\cite{Moulavi2014DensityBasedCV}. Lastly, we evaluate how robust our approach is by measuring the alignment between two runs with different initializations by CBA from Eq.~\ref{eq:CBA}.
\begin{figure}
    \centering
    \includegraphics[width=\linewidth]{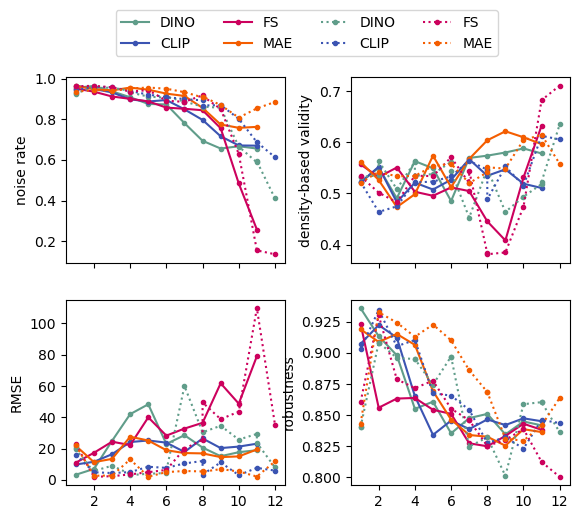}
    \caption{We evaluate the quality of concept discovery. RMSE measures the MSE between the distance matrix of the original and embedded activations and shows how faithfully the UMAP embedding captures the geometry of the representation. DBCV is a density-based clustering validity index that contrasts intra- vs inter-cluster density. The noise rate is the ratio of points classified as noise in HDBSCAN. Robustness is measured between two runs by concept-alignment from \cref{eq:CBA}. Results are across layers for CLS (dotted) and SEQ (solid) token representations of the models in \cref{tab:models}.  }
    \label{fig:concept_discovery}
\end{figure}
Before discussing the results on embedding and clustering quality (see Fig.~\ref{fig:concept_discovery}), we detail the hyperparameter tuning process for UMAP and HDBSCAN.
For UMAP, we tune the minimal distance parameter to enhance local cluster density, acknowledging that a lower minimal distance can increase noise. The number of neighbors parameter controls the local structure captured by the embedding; smaller values capture finer local neighborhoods but may distort the global structure, which is important for subsequent concept alignment analysis. We also experiment with the embedding dimensionality, constrained by practical considerations—the curse of dimensionality renders density clustering in the original high-dimensional representation infeasible, where $F'=50$ is the practical limit for the embedding dimensionality.
For HDBSCAN, the minimum cluster size parameter is tuned to balance between identifying noise and merging distinct clusters; a too-small value may recognize noise as clusters, while a too-large value could merge distinct clusters. We set the min samples parameter, which controls the algorithm's conservativeness regarding noise, relatively low due to sampling limitations - some concept manifolds may not be sampled densely enough. We tune all these hyperparameters to maximize the DBCV across models and layers. During hyperparameter tuning, we weight the average DBCV across clusters by their respective sizes to indirectly account for the noise rate. The final hyperparameters used in all subsequent experiments are reported in the appendix.
Turning to the results presented in Fig.~\ref{fig:concept_discovery}, we observe that
RMSE increases slightly across layer for most models. Only for FS there is a strong increase from layer eight onwards, indicating these representations are more difficult to embed and we can trust the clustering on the embedding less which has a high DBCV but low robustness.
The density-based validity is medium, but similar across models and SEQ tokens vs. CLS tokens. Given how challenging the clustering task is, we view this result as decent and refer to the convincing qualitative impression of the clusters in \cref{fig:concept_formation} and \cref{fig:concept_atlas}.
Noise rates are rather high but decrease across layers. 
The high noise may be due to insufficiently dense sampling, i.e. thorough sampling of noisy regions could result in concept clusters. However, the number of input samples is restricted computationally by UMAP and HDBSCAN.
Robustness decreases for all models across layers but stagnates at around $0.84$ for most models in the late layers. This links back to the trend in RMSE which shows that higher layers are harder to embed.
For the qualitative evaluation of our concept discovery method, we construct \emph{concept formation graphs} (CFGs) that depict the flow of token assignments to concepts from one layer to the next as an unweighted, directed graph. \cref{fig:concept_formation} displays the formation of the ``apples'' concept throughout the layers of the FS model. Note that these graphs may be incomplete, as some nodes might not be detected by the clustering method, illustrating the under-sampling problem described above. Additional examples for other models and the detailed algorithm for CFG construction are provided in the appendix.
\subsection{Sanity checking concept structure for alignment} \label{sec:sanity_check}
We use a sanity check to demonstrate how concept-based alignment analysis benefits from concepts defined as non-linear manifolds by comparing against concept alignment based on other definitions and discovery methods. The sanity check is based on the assumption that neighboring representations should be most aligned. We measure the ratio of layers for which a neighboring layer is most aligned under CBA from \cref{eq:CBA}  
We compare NLMCD concepts against one-dimensional linear subspaces discovered by~\cite{fel2023holistic}, multi-dimensional linear subspaces discovered by MCD~\cite{Vielhaben2023}, and spherical concepts discovered by KMeans clustering~\cite{fel2023holistic}. To obtain soft concept membership scores for the linear subspaces, we project the feature vector onto the concept subspace and clip to negative values to 0, as we argue that a feature vector pointing into the opposite direction of a concept signifies the concept not being active. For KMeans concepts, we measure concept proximity by the euclidean distance to the cluster centroid. We also normalize concept membership scores $P^{\alpha'} =  P^{\alpha} / \sum_{\alpha} P^{\alpha}$ as their sum is required to be less bounded by one  $\sum_{\alpha} P^{\alpha} \leq 1$ in Eq.~\ref{eq:CBA_dist}. There is no direct way to estimate the number of concepts for PCA, MCD and KMeans, so we use all $F=768$ components for PCA for a conservative baseline, and the number of concepts discovered by NLMCD for MCD and KMeans discovery.
\begin{table}[t]
    \centering
\begin{tabular}{llrrrr}
\toprule
 &  & FS & CLIP & DINO & MAE \\
\midrule
SEQ  & PCA & 0.45 & 0.90 & 1. & 0.60 \\
 & MCD & 0.81 & 0.90 & 0.81 & 0.70 \\
& KMeans & \textbf{0.90} & 0.90 & 0.81 & \textbf{1.} \\
 & \textbf{NLMCD} & \textbf{0.90} & \textbf{1.} & \textbf{0.90} & \textbf{1.} \\
\cmidrule{1-6}
 CLS& PCA & 0.75 & 0.72 & 0.50 & 0.33 \\
  &MCD  & 0.33 & 0.16 & 0.16 & 0.16 \\
 & KMeans & 0.83 & 0.91 & 0.50 & 0.66 \\
 & \textbf{NLMCD} & \textbf{1.} & \textbf{1.} & \textbf{0.83} & \textbf{0.75} \\
\bottomrule
\end{tabular}
    \caption{Sanity check for concept alignment, showing the percentage of layers where neighboring layers exhibit the highest alignment across various ViT models. We compare the suitability of  NLMCD concepts against other methods: one-dimensional linear subspaces (PCA), multi-dimensional linear subspaces (MCD), and spherical concepts (KMeans). NLMCD consistently outperforms other approaches, particularly in CLS token alignment, with the exception of DINO SEQ where PCA achieves the highest score.}
    \label{tab:sanity check}
\end{table}
We present the scores in Tab.~\ref{tab:sanity check} for SEQ and CLS token concept alignment. We find that our approach performs best across all models except DINO where PCA achieves the highest score. All other concept frameworks reach NLMCD scores only for single models. For the CLS token, the gap between NLCMD and the other methods is larger than for SEQ token alignment.
\subsection{Concept Alignment Analysis} \label{sec:results_alignment}
We now investigate concept-based alignment described in Sec.~\ref{sec:methods_alignment} between representations across layers and models. We structure the analysis into \emph{intra-model} and \emph{inter-model}. Due to limited space, we focus on SEQ representation and defer the CLS representation analysis to the appendix.
\begin{figure*}
    \centering
    \includegraphics[width=\textwidth]{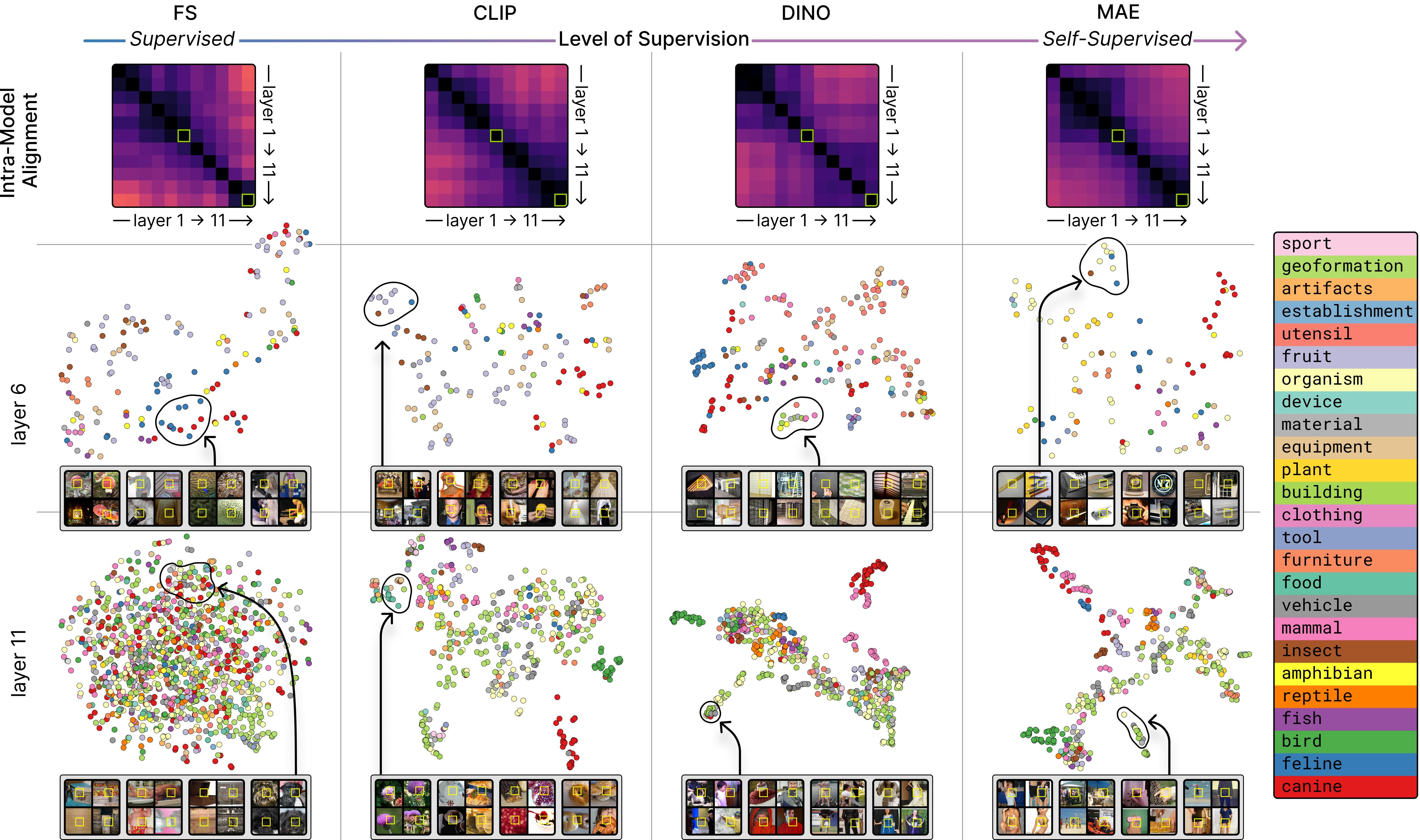}
    \caption{Intra-model relationships based on SEQ representations across layers. In the \textbf{upper row}, we show CBA from \cref{eq:CBA} to visualize how representations are transformed across layers of the models from \cref{tab:models} (darker pixels correspond to higher alignment). We observe a nucleation process between layer 9 and 10 in FS and smoother processing split into two major blocks between layer 1-6 and 6-11 in CLIP, DINO and FS. In the \textbf{center and bottom row} we zoom into the representations at layer 6 and 11 of each model and partition the scalar CBA alignment into single concepts. We show a UMAP embedding constructed from the pairwise distance of concept measured by $d_{cross}(P^{\alpha},P^{\beta} )$ from \cref{eq:CBA_dist_c}. Each point in this \emph{concept atlas} corresponds to a distinct concept $P^{\alpha}$. To convey their meaning, we show four random input tokens from the members of the concept cluster $P^{\alpha}$ marked by a yellow box in the entire image. The higher the level of supervision of ViT training ranging from FS, over CLIP to DINO and MAE, the less semantically organized are the representations at layer 11.}
    \label{fig:concept_atlas}
\end{figure*}
\vspace{-0.3cm}
\paragraph{Intra-model representations} We analyze how representations are transformed within one model and how they are structured across layers. To supplement concept-based alignment analysis between representations, we further evaluate alignment with labels from ImageNet-1k and token location and the intrinsic dimensionality of each concept.
With this analysis, we answer the questions: 1) Where does the model representation change the most and how? 2) Which concepts are encoded in lower layers vs. upper layers? 3) How structured are the latent representations - does the model encode semantically similar concepts in spatial proximity to each other? opened in the introduction. 
\begin{figure}
    \centering
    \includegraphics[width=1.0\linewidth]{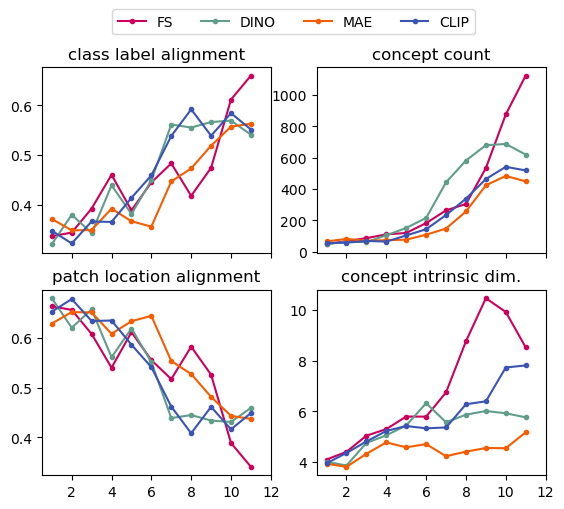}
    \caption{Class label alignment, token location alignment (both based on CBA from \cref{eq:CBA}), concept count, and the average intrinsic dimensionality (based on~\cite{Facco2017}) across concepts supplement the intra-model alignment analysis, by providing insights into how well the model aligns with ImageNet-1k labels, the spatial organization of tokens, and the complexity of the learned concepts as they evolve through the layers.}
    \label{fig:model-insights}
\end{figure}

\paragraph{Where does the model representation change the most and how?}
First, we focus on the intra-model alignment heatmaps between SEQ representations across layers measured by CBA from \cref{eq:CBA} in the upper row of Fig.~\ref{fig:concept_atlas}. 
Interestingly, the transformation process in CLIP, DINO and MAE models is split between the first, i.e., layer one to six, and the second model half, i.e., layer six to eleven. 
The concept characteristics in Fig.~\ref{fig:model-insights} reflect this break and give insight into how the representation is transformed between the break from layers six and seven. The concept count increases rather smoothly across layers for these models, but picks up at layer seven.  In contrast, class alignment has a marked increase at this point. For DINO and MAE, the average intrinsic dimensionality of concepts slightly decreases at this point but increases further for CLIP. Lastly, token location alignment also has the most significant decrease at this point.
In contrast to the models above, the FS model exhibits a pronounced change rate between nine and ten, resulting in a sudden enhancement in class alignment at layer 10, accompanied by a marked increase in the number of clusters and intrinsic dimensionality. This is reflected in a low alignment between representations in the last two blocks of the FS model and indicates a nucleation process, where concepts begin to separate into distinct classes used for supervised training. This nucleation process has been previously observed in ResNets~\cite{Doimo2020}.

\paragraph{Which concepts are encoded in lower layers vs. upper layers? How structured are the representations?} We now zoom in and partition the representation into single concepts, at layer six just before the block separation in CLIP, DINO, MAE and at layer eleven as the last layer of the second model part.
We construct a UMAP embedding based on the distance between concept pairs measured by $d_{cross}(P^{\alpha},P^{\beta} )$ from \cref{eq:CBA_dist_c}. Each point in this \emph{concept atlas} corresponds to a different concept $P^{\alpha}$. To convey their meaning, we show four random input tokens from the members of the concept cluster $P^{\alpha}$ (framed by a yellow box). Concept atlases for the representation at layer six and eleven across all models in \label{fig:qualitative_alignment_CLIP}  give a visual impression of how semantically organized the concepts are. To guide the eye, we color-code the concept clusters based on categories derived from the ImageNet-1k labels of the images from which the patches were extracted. We first map these labels to more abstract categories\footnote{The mapping is provided in the appendix. Note that this labeling is only a proxy and may not accurately reflect the actual content of the patches—for instance, a patch might show grass on which an animal stands.} using the WordNet hierarchy~\cite{miller1995Wordnet}.  After mapping, we perform a majority vote among all patches in a cluster to assign the category.
At layer six, concepts appear structured, but not yet aligned with the WordNet categories. By visual inspection, concepts are less abstract and rather encode structures, shapes and object parts. Representations across ViTs at this layer show a similar level of structuredness.
In contrast, at layer eleven, the FS representation is notably less semantically organized than that of the other models. For CLIP, DINO, and MAE we point out how well the canine concepts are separated. To further exemplify, human body parts like neck, shoulder, and legs are grouped together in the representation of DINO and MAE. This alignment requires not only the preservation of local, or intra-cluster distances, but also the maintenance of broader, inter-cluster distances. 
We conclude that supervised training for the FS model does not enforce this level of semantical organization. In fact, it might make sense to push similar concepts apart in feature space to avoid confusion. However, this likely has negative implications for generalization to other tasks.
\begin{figure}
    \centering
    \includegraphics[width=\linewidth]{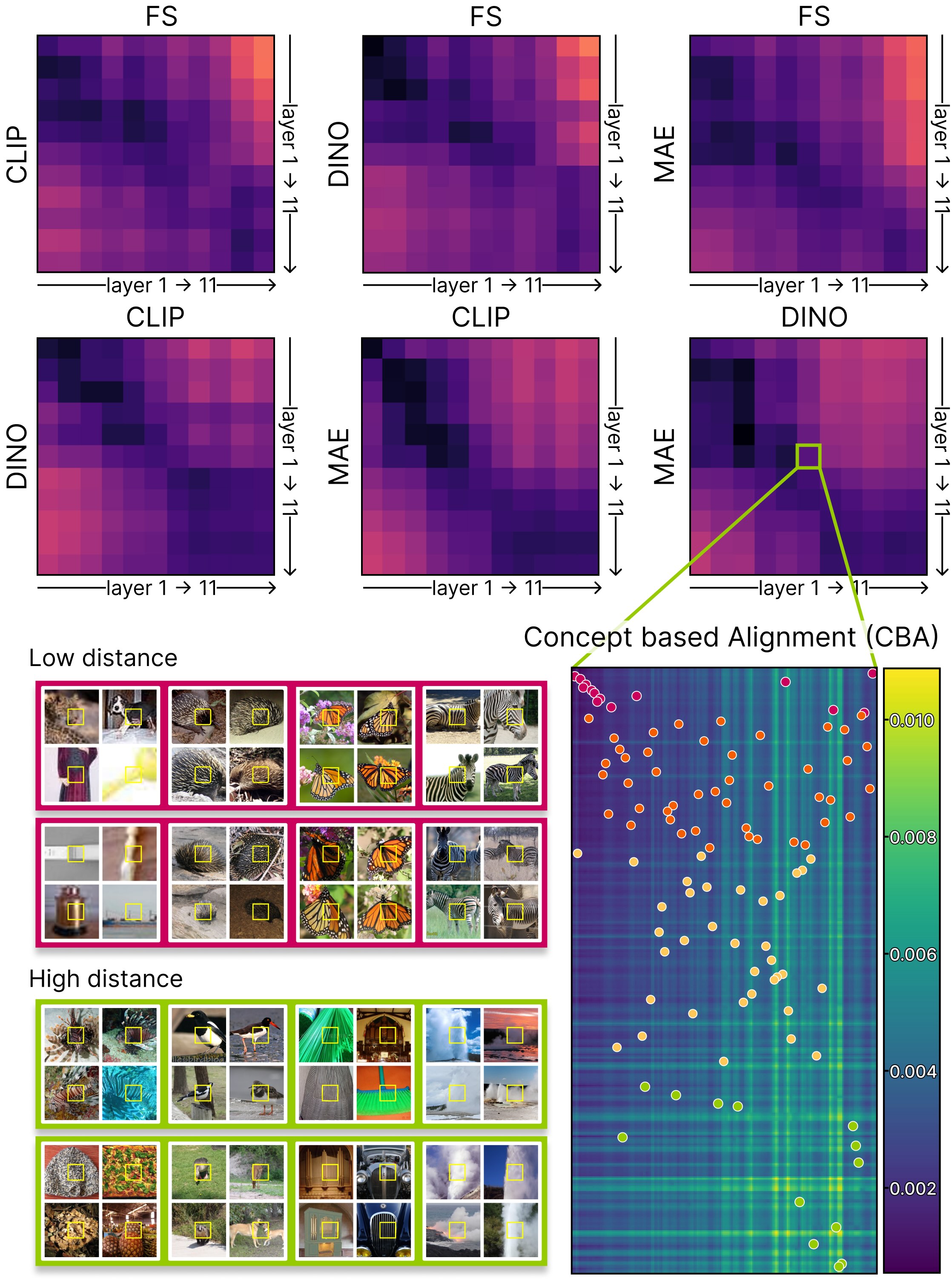}
    \caption{ Inter-model relationships based on SEQ representations across layers. In the \textbf{upper part}, we show CBA from \cref{eq:CBA} to visualize how representations differ between the models in \cref{tab:models}  (darker pixels correspond to higher alignment). In the \textbf{lower part} we zoom in into the concept-wise distances $d_{cross}(P^{\alpha},P^{\beta} )$ from \cref{eq:CBA_dist_c} between the representation of layer six in MAE and DINO. We give examples of universal concepts with low distance and unique ones with high distance from matches of pairs that minimize the total distance.}
    \label{fig:inter-model}
\end{figure}
\vspace{-0.1cm}
\paragraph{Inter-model relations}
We now analyze how the representations between two different models differ and present $CBA$ from \cref{eq:CBA} between all layers of the models from \cref{tab:models} in the upper part of \cref{fig:inter-model}.
We observe higher alignment between the self-supervised models DINO and MAE than with CLIP and the FS model in the alignment heatmaps. Further, layers of the first are more aligned than those of the second half across all models pairs. We conclude that basic foundational features are learned similarly across models, while later layers diverge as the models specialize to concepts serving their pre-training task.

\paragraph{How is the representation of model A different from that of model B?}
We zoom in into the distance $d_{cross}(P^{\alpha},P^{\beta} )$ from \cref{eq:CBA_dist_c} between concept pairs from representations of DINO and MAE layer six in the lower part of \cref{fig:inter-model}. We visualize the distance matrix between all concepts. To select specific examples of pairs for inspection, we match concepts via the Hungarian algorithm that minimizes the sum of distances of pairs. From this selection, we show concept pairs with low distance (blurriness, satchel of a hedgehog, zebra stripes) 
and high distance(complex high-frequency structure, vertically textured structure, fountain and fog-like). Both the universal concepts with low distance and the more specific concepts with high distance seem to correspond mainly to structure and texture but visual discrepancy is more pronounced for high-distance concepts.

\section{Conclusion}
We propose a novel approach that combines concept discovery with representational alignment analysis in ViTs.
With concept-based alignment analysis, we answer the questions raised in the introduction and examine the structuredness in feature spaces of different ViTS, as well as fine details between the concepts of two different models. These insights are not available through traditional scalar alignment measures. 
Understanding the structured nature of latent spaces can guide practitioners in choosing models that not only perform well on benchmark datasets but also exhibit robust feature representations for downstream tasks. For instance, the nucleation process in FS emphasizes the importance of model structure over mere classification accuracy when selecting a pre-trained model.

\paragraph{Limitations}
The computational scalability of HDBSCAN limits the sampling of feature vectors which makes undersampled concept regions appear as noise. The limited variability of ImageNet-1k might obfuscate the meaning of a concept, e.g. when a concept represents a color but there are only dog patches of that color.

\paragraph{Acknowledgments} This work was supported by the European Union’s Horizon Europe research and innovation programme (EU Horizon Europe) as grants [ACHILLES (101189689), TEMA (101093003)]; and the German Research Foundation (DFG) as research unit DeSBi [KI-FOR 5363] (project-id: 459422098).

{
    \small
    \bibliographystyle{ieeenat_fullname}
    \bibliography{main}
}

\appendix

\section{HDBSCAN}

After concept discovery with HDBSCAN, we compute concept proximity scores $\clustering{P}_{\{\phi\}}= \{\clustering{P}(\phi_0), \dots, \clustering{P}(\phi_N)\}$, where $\clustering{P} \in [0,1]^n$ holds the concept proximity scores $P{\alpha}$ of each concept $C^{\alpha}$. These rely on the implementation of soft clustering with HDBSCAN from  \cite{McInnes2017}, which we formalize here for the reader's convenience. 

\paragraph{Clustering} 
HDBSCAN first transforms the feature space using a density-informed metric called \emph{mutual reachability distance}
\begin{align}
    \text{MRD}(\phi_i,\phi_j) = max(&\text{coreDistance}_k(\phi_i), \nonumber \\ &\text{coreDistance}_k(\phi_j),d(\phi_i,\phi_j))
\end{align}
where $\text{coreDistance}_k(\phi)$ is the distance between a point $\phi$ and its $k$-nearest neighbor.
Based on the mutual reachability distance between all pairs, a minimum spanning tree is constructed that connects all points and minimizes the sum of the edges weighted by MRD. From this, a hierarchical tree is constructed via robust single linkage clustering. 
The hierarchical tree is condensed by eliminating insignificant clusters and simplifying the hierarchy. This is achieved by selecting a range of \emph{persistence} values $\lambda$, which are the inverses of the mutual reachability distances ($\lambda = 1 / \text{MRD}$). Clusters that persist over significant ranges of $\lambda$, i.e. they are stable across multiple density levels, are retained, while clusters that exist only over narrow ranges of $\lambda$ are considered noise and pruned from the tree. The result is a condensed tree that focuses on the most significant clusters.
Finally clusters are extracted from the condensed tree either based on their stability across different density levels or simply the leaf nodes are identified as clusters.

\paragraph{Soft clustering with HDBSCAN}
The soft cluster membership scores combine a distance-based membership with and an outlier score.

For the \textbf{distance-based membership} to cluster  $C^{\alpha}$, first $k$ exemplar points $\{\phi^{\alpha}_i\}$, $i\in[1,k]$, are extracted. A single centroid is not enough to characterize a cluster as its shape can be arbitrary. The exemplar points are the points within the leaf nodes beneath cluster $C^{\alpha}$ with maximum persistence  $\lambda $ in the condensed tree, i.e. the densest points where the cluster persists.

Then, the distance membership score between a point $\phi$  and a cluster $C^{\alpha}$ is the inverse minimum distance across the exemplar points $\{\phi^{\alpha}_i\}$,

\begin{equation}
    M^\alpha(\phi)_{\text{dist}} = \frac{   [\text{min}_i (d(\phi, \phi^{\alpha}_i))]^{-1} }{  \sum_{\beta}  [\text{min}_j( d(\phi, \phi^{\beta}_j))]^{-1} } \, ,
\end{equation}
normalized across all clusters.

The \textbf{outlier-based membership} compares a point's membership persistence to the total persistence of a cluster:

\begin{equation}
M^{\alpha}(\phi)_{\text{membership}} =  \frac{ \lambda_{\phi \rightarrow C^\alpha} - \lambda_{\text{birth}}^{C^\alpha} }{ \lambda_{\text{max}}^{C^\alpha} - \lambda_{\text{birth}}^{C^\alpha} } \, .
\end{equation}
Here, $ \lambda_{\text{birth}}^{C^\alpha} $ is the persistence value at which cluster $ C^\alpha $ first appears, i.e. its birth point in the condensed tree and $ \lambda_{\phi \rightarrow C^\alpha} $ is the persistence value at which point $\phi$ would join cluster $C^\alpha$. 

Finally, distance and outlier-based membership are combined with stronger emphasis on outlier-based membership,
\begin{equation}
    M^{\alpha}(\phi) = (M^{\alpha}(\phi)_{\text{dist}})^{1/2} \cdot (M^{\alpha}(\phi)_{\text{membership}})^2 \, ,
\end{equation}

and normalized $M_{\text{norm}}^{\alpha}(\phi) =  M^{\alpha}(\phi) / \sum_{\beta} M^{\beta}(\phi) $. 

This membership score  $M_{\text{norm}}^{\alpha}(\phi)$ can be interpreted as the probability that a point $\phi$ belongs to cluster $C^{\alpha}$,   given that the point belongs to some cluster,
\begin{equation}    
 M_{\text{norm}}^{\alpha}(\phi) \equiv P(\phi \in C^\alpha \mid \exists \beta : \phi \in C^\beta) \, .
\end{equation}

We want to compute the joint probability $P(\phi \in C^\alpha)$, which includes the probability that $\phi$ may be noise,

\begin{equation} 
P(\phi \in C^\alpha) = P(\phi \in C^\alpha \mid \exists \beta : \phi \in C^\beta)  P(\exists \beta : \phi \in C^\beta) \, .
\end{equation}

Here, $P(\exists \beta : \phi \in C^\beta)$ is the probability that $\phi$ belongs to some cluster.
To estimate $P(\exists \beta : \phi \in C^\beta)$, the $\lambda$ value at which $\phi$ would join the nearest cluster is compared to the maximum $\lambda$ value of that cluster,

\begin{equation} P(\exists \beta : \phi \in C^\beta) = \frac{\lambda_{\phi \rightarrow C^\alpha}}{\lambda_{\text{max}}^{C^\alpha}} \, ,
\end{equation}

where $\lambda_{\phi \rightarrow C^\alpha}$ is the persistence value at which point $\phi$ would join its nearest cluster $C^\alpha$ and $\lambda_{\text{max}}^{C^\alpha}$ is the maximum $\lambda$ value of cluster $C^\alpha$.
Thus, the final probability, that point $\phi$ belongs to cluster $C^{\alpha}$ is,
\begin{equation}
    P^{\alpha}(\phi) =  \frac{\lambda_{\phi \rightarrow C^\alpha}}{\lambda_{\text{max}}^{C^\alpha}} \cdot M_{\text{norm}}^{\alpha}(\phi)  \, . 
\end{equation}

\section{Details on experimental setup}
Here, we provide further details on the experiments. 

\paragraph{ViT sources} We list the URL of each Vision Transformer provided by the timm library \cite{rw2019timm}: 
\begin{itemize}
    \item FS: \url{https://huggingface.co/timm/vit_base_patch16_224.augreg_in1k}
    \item CLIP: \url{https://huggingface.co/timm/vit_base_patch16_clip_224.openai}
    \item DINO: \url{https://huggingface.co/timm/vit_base_patch16_224.dino}
    \item MAE: \url{https://huggingface.co/timm/vit_base_patch16_224.mae}
\end{itemize}

\paragraph{Hyperparameters for UMAP and HDBSCAN} We tune hyperparameters of UMAP and HDBSCAN such that the density-based validity index DBCV is maximized across models and layers. Here, for DBCV, the average across clusters is weighted by their respective size such that the noise rate is indirectly included. We re-iterate the effect of the most influential hyperparameters that we tune and state the final value we used:
\begin{itemize}
    \item \textbf{Minimal distance in UMAP}: a low minimal distance in UMAP enhances local cluster density but may also increase noise. We use a value of 0.01 in all experiments.
    \item \textbf{Number of neighbours in UMAP}:  the number of neighbors controls the local structure, the smaller the finer it captures local neighborhoods but distorts global structure which is important for concept alignment analysis later. We use a value of 30 in all experiments.
    \item \textbf{Embedding dimensionality in UMAP}:  We use the practical limit for HDBSCAN of $F'=50$ in all experiments. 
    \item \textbf{Minimum cluster size in HDBSCAN}: a too small minimum cluster size may identify noise as a cluster, whereas, when too large, distinct clusters will merge. We use a value of 50 in all experiments.
    \item \textbf{Min samples in HDBSCAN}: controls how conservative the algorithm is about noise. We need this to be rather low because of sampling limitations which means that most likely some concept manifolds are not sampled densely enough. We use a value of 20 in all experiments.
\end{itemize}
Additionally, we assume that clusters are rather uniform in size and select the leaf nodes in the HDBSCAN hierarchical condensed tree as clusters. 
Sampling one pooled SEQ token (we average-pool over $4 \times 4$ tokens) or one CLS token from each representation of images within a 25\% subset of the ImageNet1-1k train set results in 315.770 feature vectors $\phi_i$ for clustering. We use the cuML \cite{raschka2020machine} versions of HDBSCAN and UMAP for computation on the GPU.

\paragraph{Cluster label in Concept Atlas}
To assign a label from the WordNet Hierarchy to each concept cluster, we first assign the ImageNet-1k label of the image from which a token is extracted to its representation feature vector $\phi_i$. Then we map this to a label higher in the WordNet hierarchy by the mapping in \cref{tab:wordnet_mao}. We then assign the most frequent label among the cluster members $\{\phi^{\alpha}_j\}$ to the cluster $C^{\alpha}$.

\paragraph{Computation of alignment} Our concept-based alignment measure CBA is based on pairs of feature vectors $(\phi_i,\phi_j)$. To reduce run-time, we sub-sample 20\% of the 315.770 feature vectors before computing CBA. 

\section{Concept Formation Graphs}

The algorithm for construction of a concept formation graph is defined as follows:
\begin{enumerate}
\item \textbf{Token assignment:} We begin by assigning each token in each layer to either one or more concepts or marking it as noise. Soft assignments are thresholded.
\item \textbf{Transition matrix calculation:} Next, we compute transition matrices for each pair of consecutive layers, $(n, n+1)$. Each matrix entry represents the count of tokens transitioning from a concept in layer $n$ to a concept in layer $n+1$.
\item \textbf{Recursive graph construction:} The CFG for a target node is then constructed recursively. Starting from the target node, we add all predecessor concepts whose “contribution” (the proportion of incoming transitions) surpasses a specified threshold. The resulting CFG is a binary, unidirectional graph in which nodes representing noise are excluded.
\end{enumerate}

\begin{figure}
    \centering
    \includegraphics[width=\linewidth]{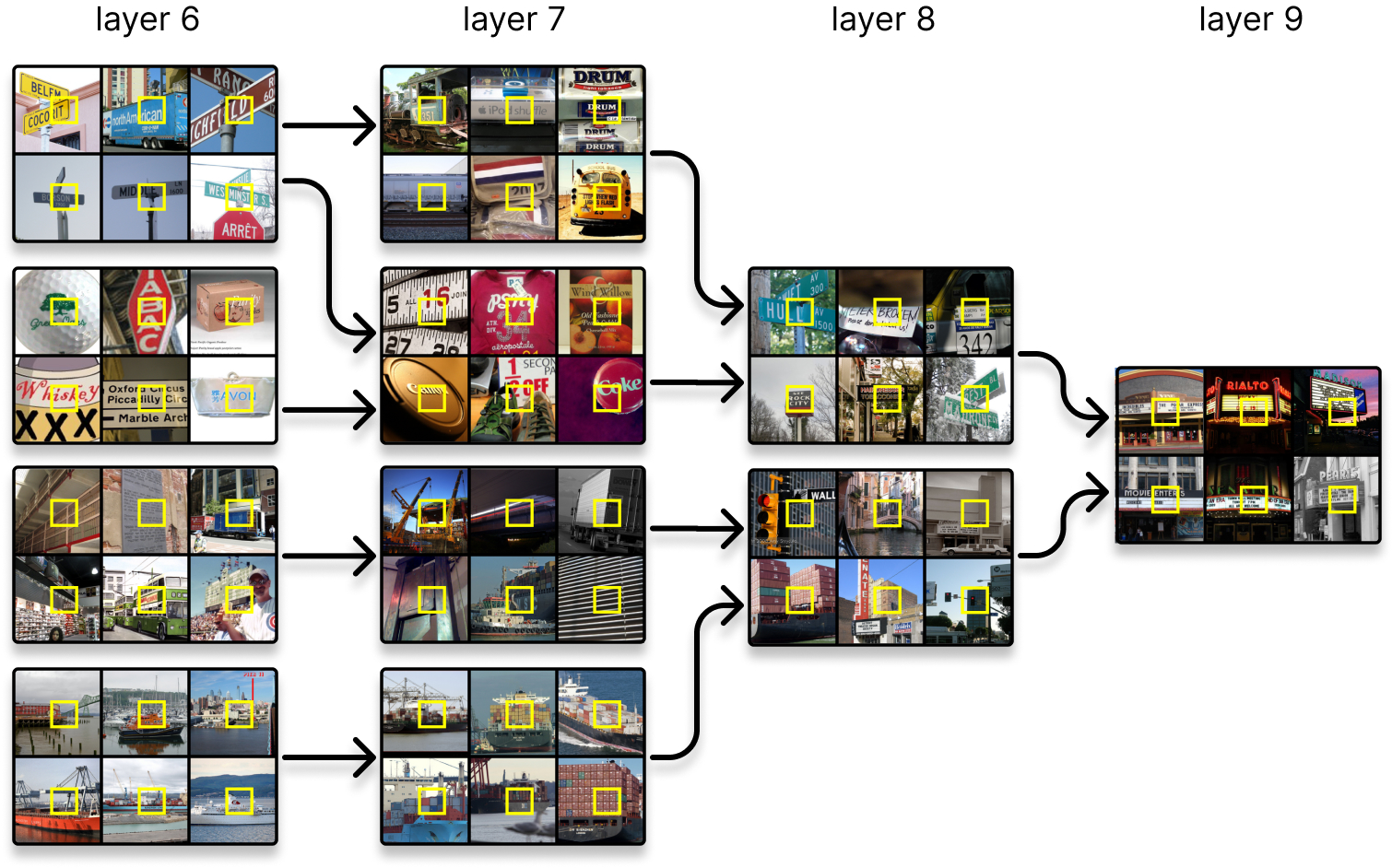}
    \caption{Concept formation graph for a concept 
    in layer 9 of DINO. Each concept is represented by six randomly sampled images containing a token assigned to that concept  (highlighted in a yellow frame).}
    \label{fig:dino_cfg}
\end{figure}

\begin{figure}
    \centering
    \includegraphics[width=\linewidth]{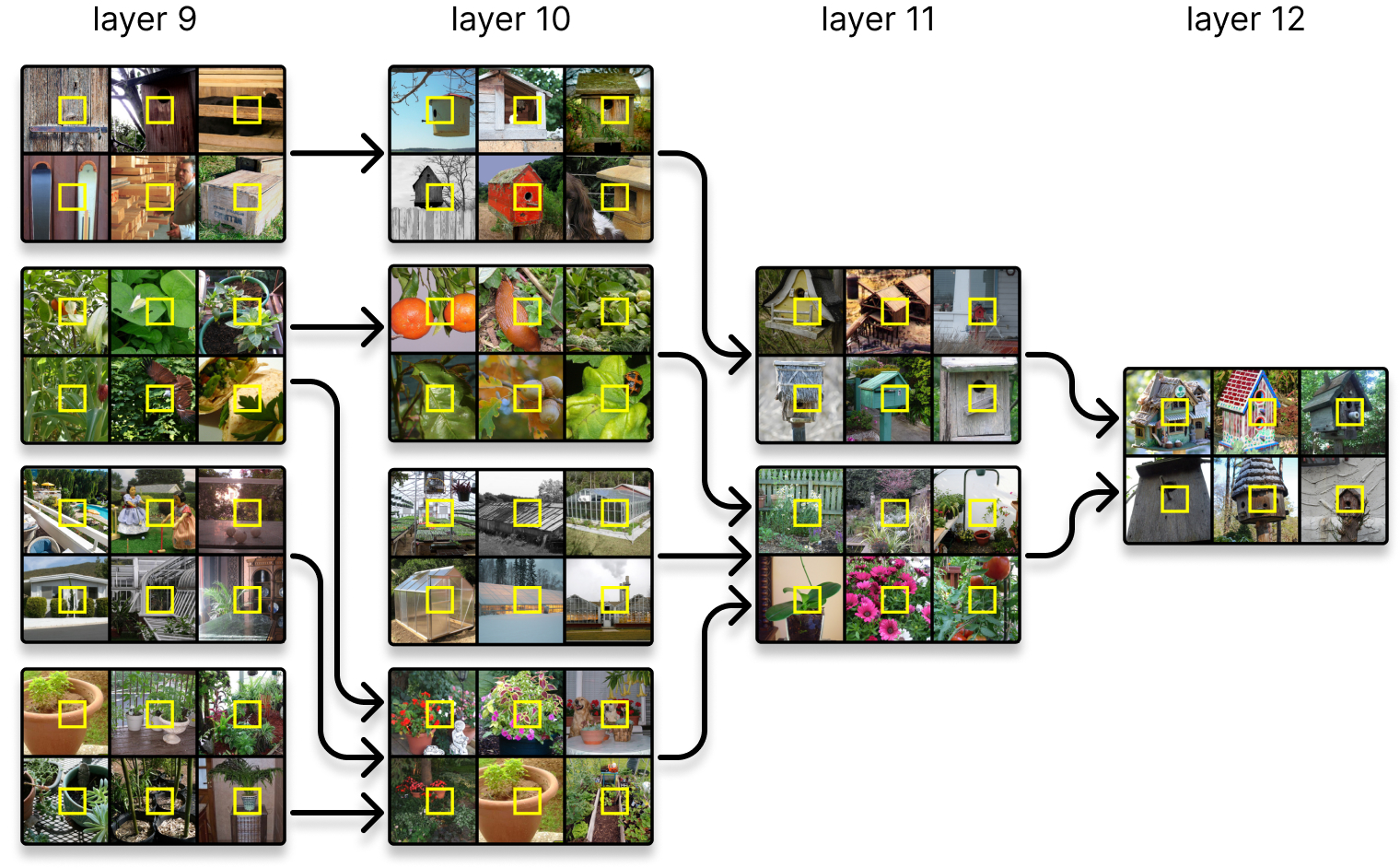}
    \caption{Concept formation graph for a concept 
    in layer 12 of CLIP. Each concept is represented by six randomly sampled images containing a token assigned to that concept  (highlighted in a yellow frame).}
    \label{fig:clip_cfg_1}
\end{figure}

\begin{figure}
    \centering
    \includegraphics[width=\linewidth]{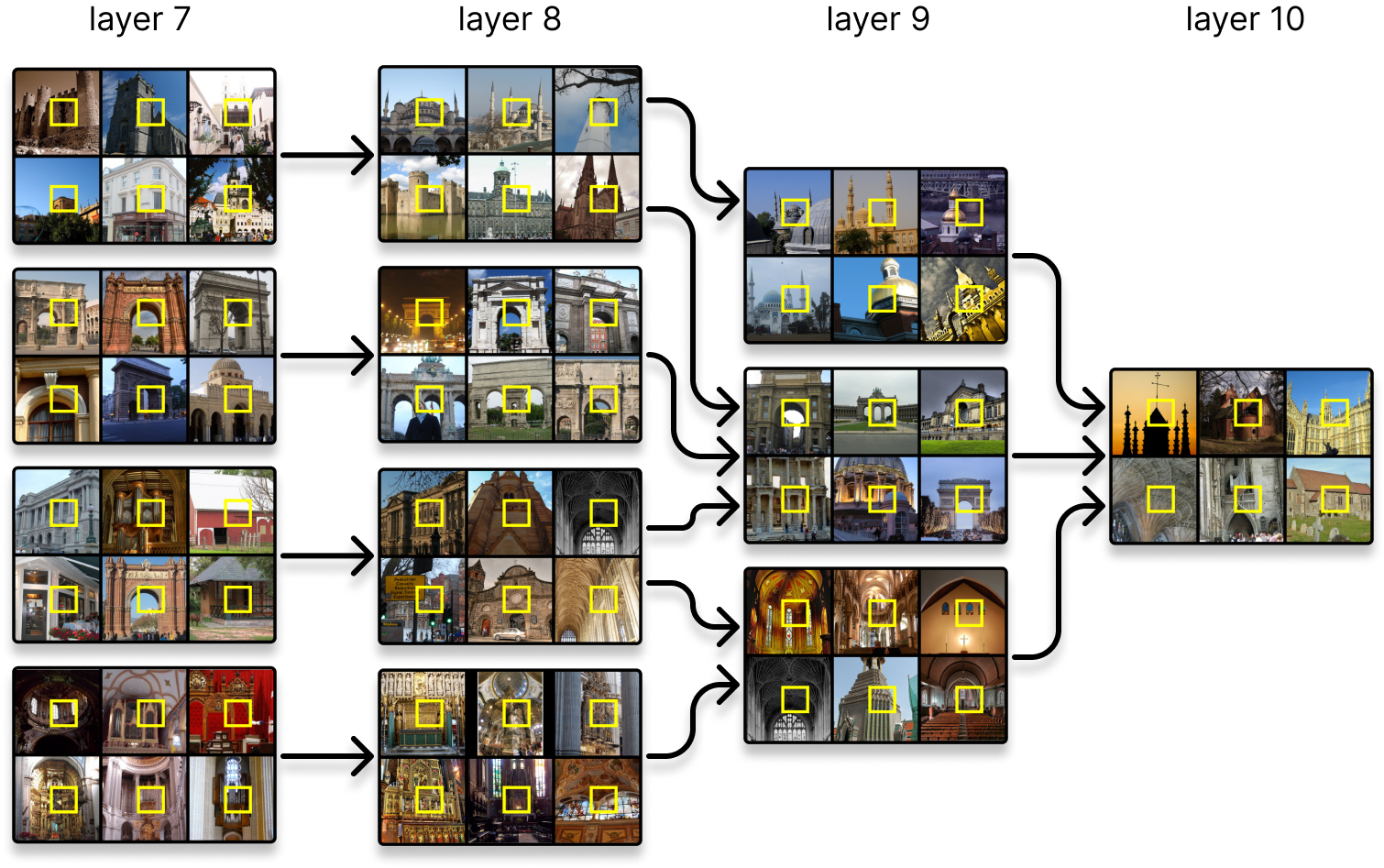}
    \caption{Concept formation graph for a concept 
    in layer 10 of CLIP. Each concept is represented by six randomly sampled images containing a token assigned to that concept  (highlighted in a yellow frame).}
    \label{fig:clip_cfg_2}
\end{figure}

\cref{fig:clip_cfg_1,fig:clip_cfg_2,fig:dino_cfg} illustrate additional exemplary CFGs for CLIP and DINO.

\section{Concept alignment analysis}

\subsection{CLS representations}

\begin{figure*}[!ht]
    \centering
    \includegraphics[width=\textwidth]{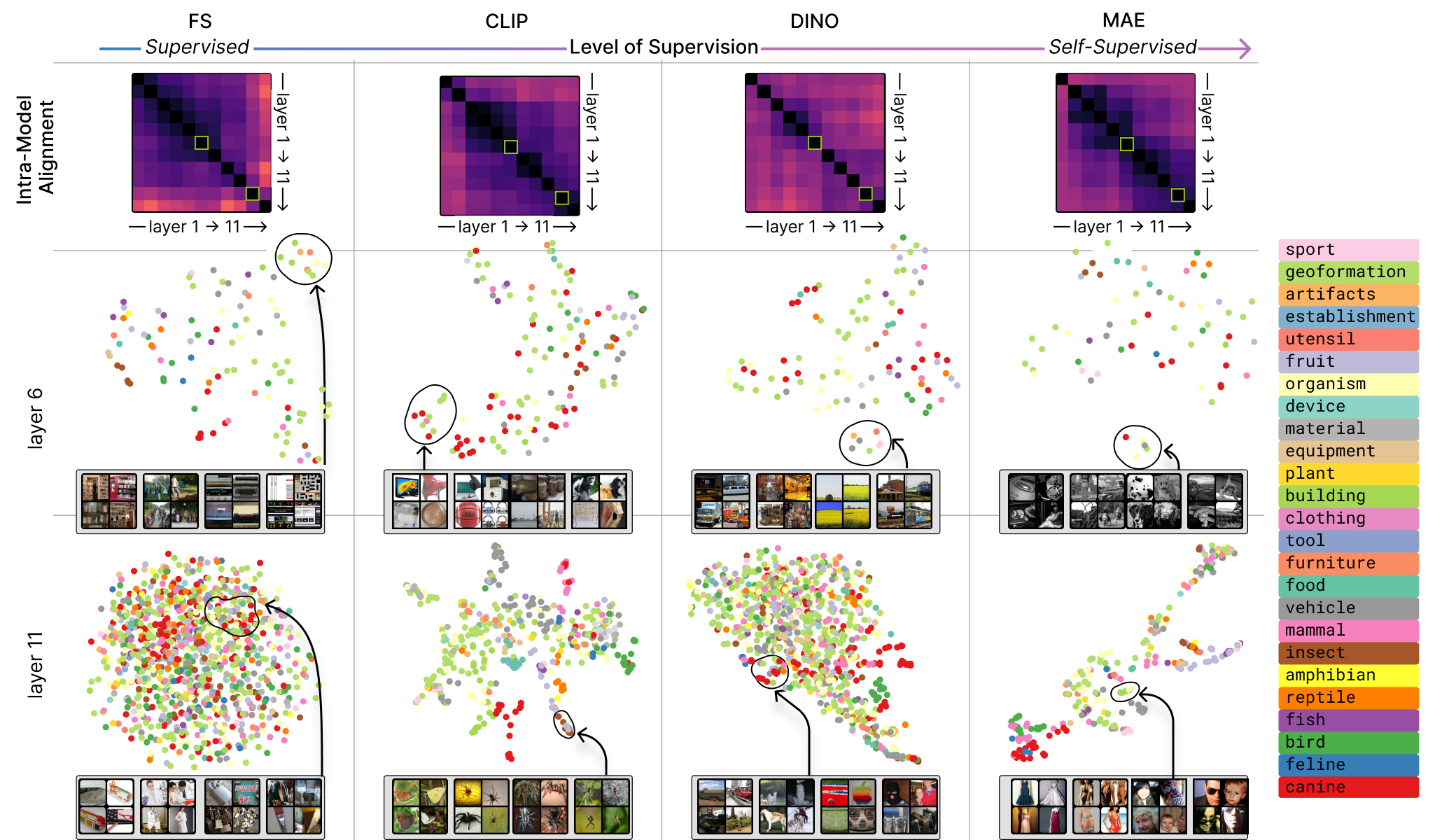}
    \caption{Intra-model relationships based on CLS representations across layers. In the \textbf{upper row}, we show CBA  to visualize how representations are transformed across layers of the models  (darker pixels correspond to higher alignment). In the \textbf{center and bottom row} we zoom into the representations at layer 6 and 11 of each model and partition the scalar CBA alignment into single concepts. We show a UMAP embedding constructed from the pairwise distance of concept measured by $d_{cross}(P^{\alpha},P^{\beta} )$. Each point in this \emph{concept atlas} corresponds to a distinct concept $P^{\alpha}$. To convey their meaning, we show four random input tokens from the members of the concept cluster $C^{\alpha}$ marked by a yellow box in the entire image.}
    \label{fig:concept_atlas_CLS}
\end{figure*}

We investigate concept-based alignment within and across models based on the CLS token representations analogous to the SEQ token analysis in the main paper.

\begin{figure}[!ht]
    \centering
    \includegraphics[width=1.0\linewidth]{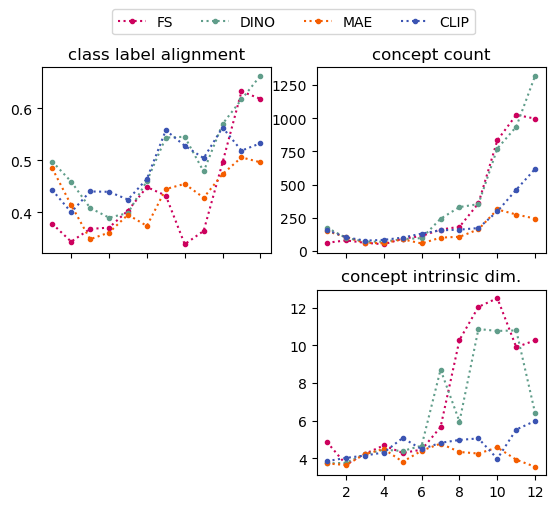}
    \caption{Class label alignment, token location alignment (both based on CBA), concept count, and the average intrinsic dimensionality (based on~\cite{Facco2017}) across concepts for CLS representations supplement the intra-model alignment analysis, by providing insights into how well the model aligns with ImageNet-1k labels, the spatial organization of tokens, and the complexity of the learned concepts as they evolve through the layers.}
    \label{fig:model-insights_CLS}
\end{figure}

\paragraph{Intra-model alignment}
First, we focus on the intra-model alignment heatmaps between CLS representations across layers measured by CBA in the upper row of \cref{fig:concept_atlas_CLS} and compare it to the same analysis between SEQ representations shown in the main paper. 
For the CLS representations of the FS model we see a very similar pattern as for the SEQ representations. Also for CLIP and MAE, the CLS intra-model alignment mirrors that of the SEQ representations; however, the first two and one blocks, respectively, show significantly lower alignment than in the SEQ tokens. This is reasonable since the model might not use these for processing information in the early blocks. Interestingly, for DINO, the CLS token alignment across layers is significantly lower than the SEQ token alignment. Class label alignment, intrinsic dimensionality of concept clusters and concept count for the CLS representations in \cref{fig:model-insights_CLS} are also similar to the SEQ results except for DINO. Here, DINO CLS concepts exhibit a notable difference to DINO SEQ concepts: the concept count, class alignment, and intrinsic dimensionality increase sharply between blocks 9 and 10 for the CLS representation but not for the SEQ representation. Lastly, the structure of the concept atlases in the lower part of \cref{fig:concept_atlas_CLS} differs the most from the structure of the SEQ concept atlases for DINO, where CLS concepts at layer 11 are less semantically organized than SEQ concepts. These observations suggest that the differences in how CLS and SEQ tokens represent and abstract information are most pronounced in DINO among the models.

\begin{figure}
    \centering
    \includegraphics[width=\linewidth]{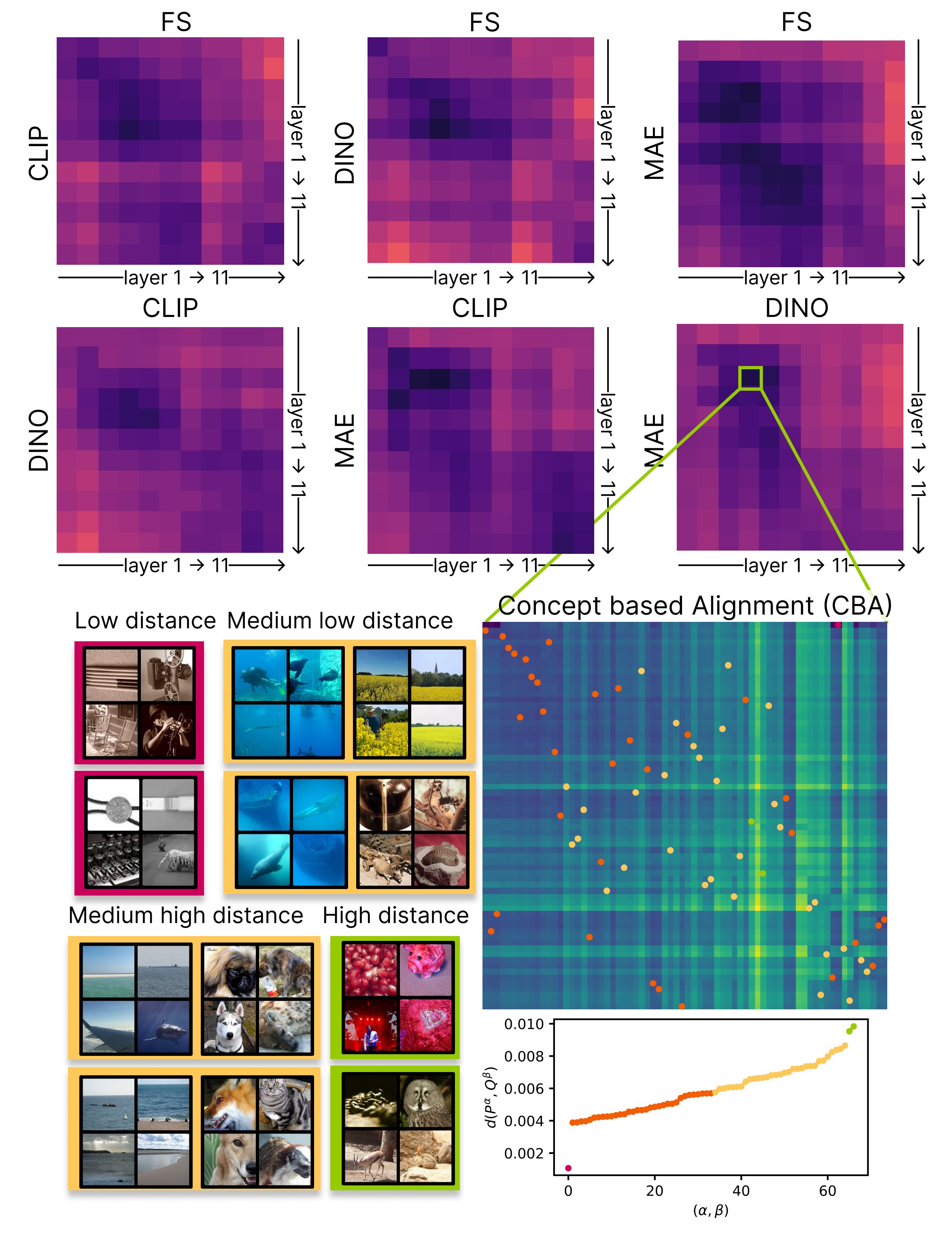}
    \caption{ Inter-model relationships based on CLS representations across layers. In the \textbf{upper part}, we show CBA  to visualize how representations differ between the models  (darker pixels correspond to higher alignment). In the \textbf{lower part} we zoom in into the concept-wise distances $d_{cross}(P^{\alpha},P^{\beta} )$ between the representation of in MAE layer 2 and DINO layer 3. We give examples of universal concepts with low distance and unique ones with high distance from matches of pairs that minimize the total distance.}
    \label{fig:inter-model_CLS1}
\end{figure}

\paragraph{Inter-model alignment} Second, we analyze how the CLS representations between two different models differ and present CBA alignment between all layers and model in the upper part of \cref{fig:inter-model_CLS1}. Like for the SEQ representations, CLS representations at layers of the first are more aligned than those of the second half across all models pairs, suggesting that basic foundational features are learned similarly across models, while later layers diverge as the models specialize to concepts serving their pre-training task. However, the overall alignment between models is weaker for CLS representations than for SEQ, also in low layers. 
Next, we zoom in into the distance $d_{cross}(P^{\alpha},P^{\beta} )$  between concept pairs from CLS representations of DINO layer 3 and MAE layer four in the lower part of \cref{fig:inter-model_CLS1}.  We visualize the distance matrix between all concepts and inspect pairs of concepts matched via the Hungarian algorithm. Most of the concepts in the pairs seem to correspond to the color composition of the images.
Visual discrepancy is more pronounced for high-distance concepts than for the other pairs with lower distance.

\subsection{Additional results for SEQ representations}

\begin{figure*}
    \centering
    \includegraphics[width=0.6\textwidth]{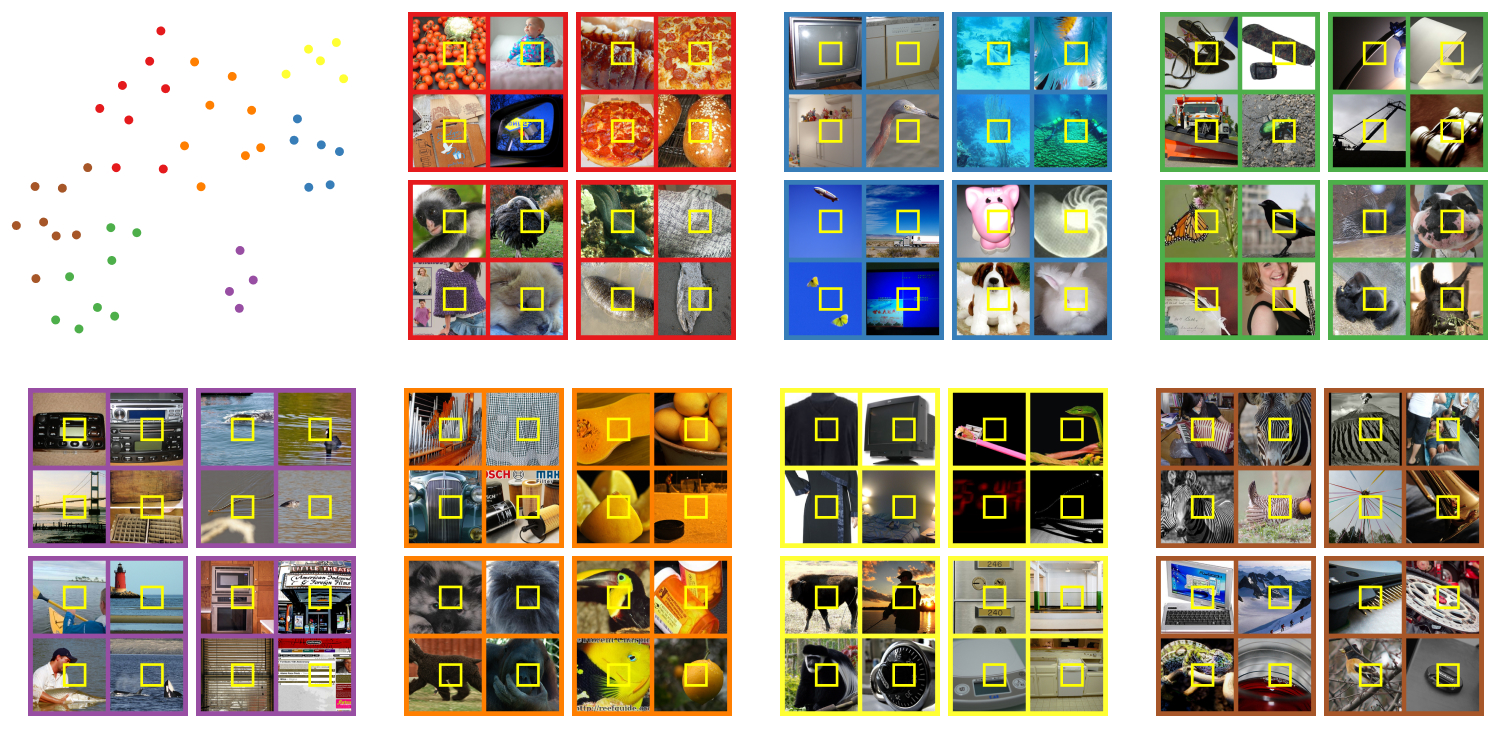}
    \caption{We zoom into the SEQ representations at DINO layer 1 and show a UMAP embedding constructed from the pairwise distance of concepts measured by $d_{cross}(P^{\alpha},P^{\beta} )$. Each point in this \emph{concept atlas} corresponds to a distinct concept $P^{\alpha}$. To convey their meaning, we show four random input tokens from the members of the concept cluster $P^{\alpha}$. We dissect the concept atlas into 7 groups and show four random concepts for each group. Concepts representing similar colors lie within the same group, e.g. shades of blue in the blue group or red and orange in the red group.}
    \label{fig:concept_atlas_DINO1}
\end{figure*}

\begin{figure*}
    \centering
    \includegraphics[width=0.85\textwidth]{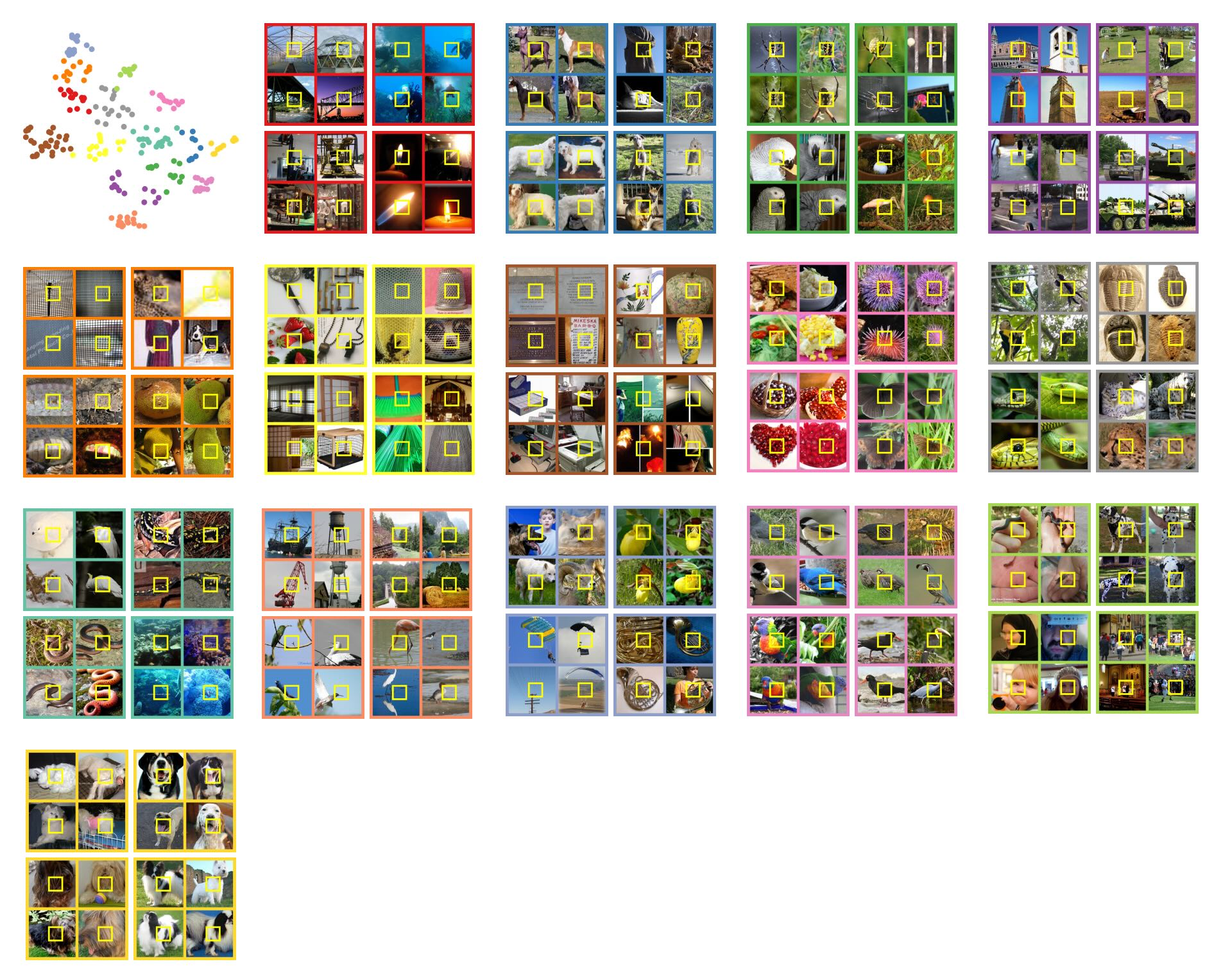}
    \caption{We zoom into the SEQ representations at DINO layer 6 and show a UMAP embedding constructed from the pairwise distance of concepts measured by $d_{cross}(P^{\alpha},P^{\beta} )$. Each point in this \emph{concept atlas} corresponds to a distinct concept $P^{\alpha}$. To convey their meaning, we show four random input tokens from the members of the concept cluster $P^{\alpha}$. We dissect the concept atlas into 15 groups and show four random concepts for each group. Most concepts represent a pattern or texture which are similar within each group.}
    \label{fig:concept_atlas_DINO6}
\end{figure*}

\begin{figure*}
    \centering
    \includegraphics[width=0.85\textwidth]{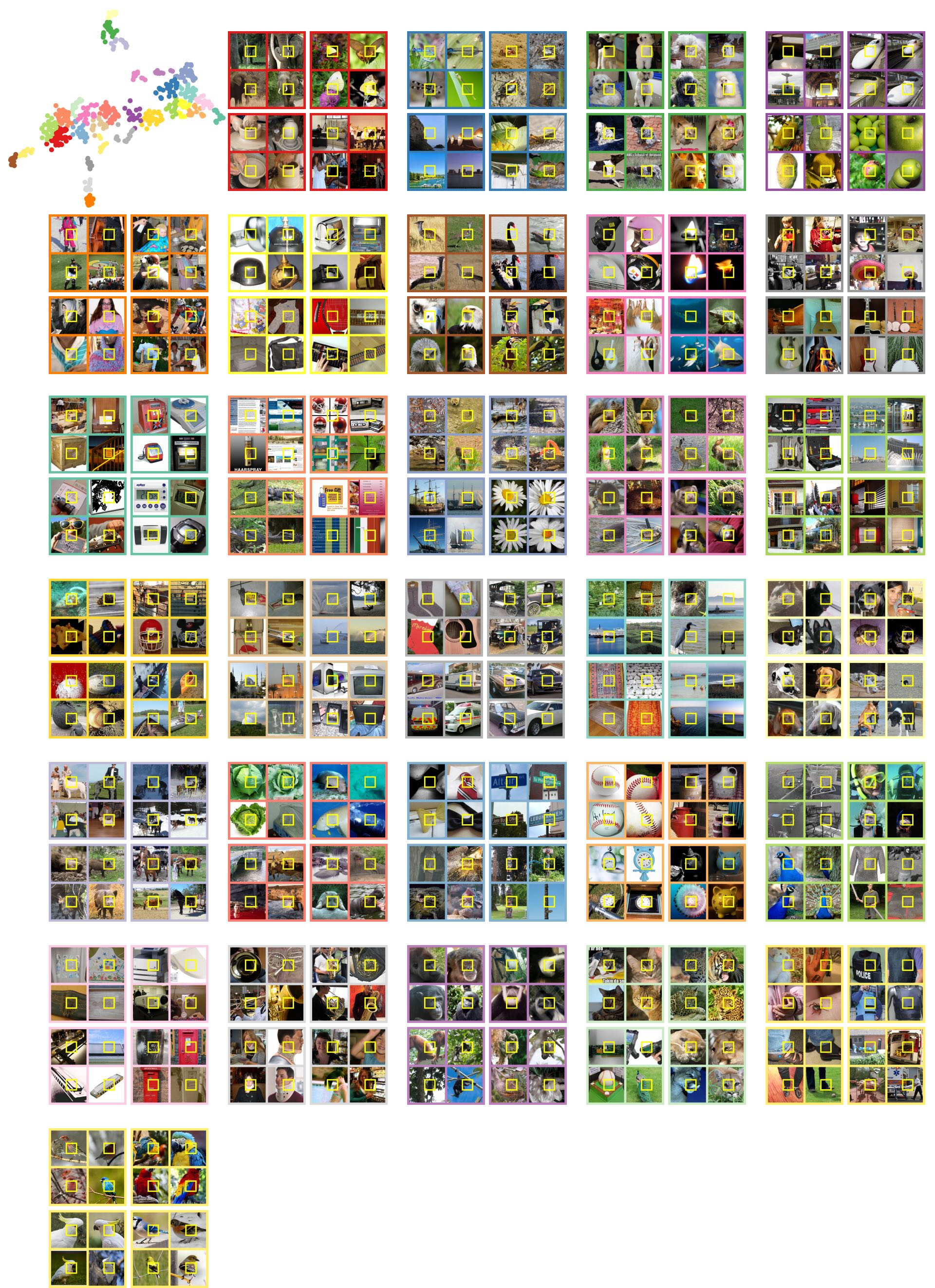}
    \caption{We zoom into the SEQ representations at DINO layer 11 and show a UMAP embedding constructed from the pairwise distance of concepts measured by $d_{cross}(P^{\alpha},P^{\beta} )$. Each point in this \emph{concept atlas} corresponds to a distinct concept $P^{\alpha}$. To convey their meaning, we show four random input tokens from the members of the concept cluster $P^{\alpha}$. We dissect the concept atlas into 30 groups and show four random concepts for each group. For most groups, these are semantically similar.}
    \label{fig:concept_atlas_DINO11}
\end{figure*}

\paragraph{Intra-model} To give a more detailed view of the organization of concepts across the layers of one model, we select the DINO model and show the respective concept atlases at layer one, six and eleven in \cref{fig:concept_atlas_DINO1}, \cref{fig:concept_atlas_DINO6}, and \cref{fig:concept_atlas_DINO11}, respectively. To give an overview of the structure within a concept atlas, we group the concepts in the UMAP embedding via KMeans and show four random concepts for each group. In layer one, many concepts correspond to color, in layer six, they represent mostly textures, and in layer eleven they correspond to abstract concepts. Moslty, concepts within a group are of similar nature.

\begin{figure}[h]
    \centering
    \includegraphics[width=\linewidth]{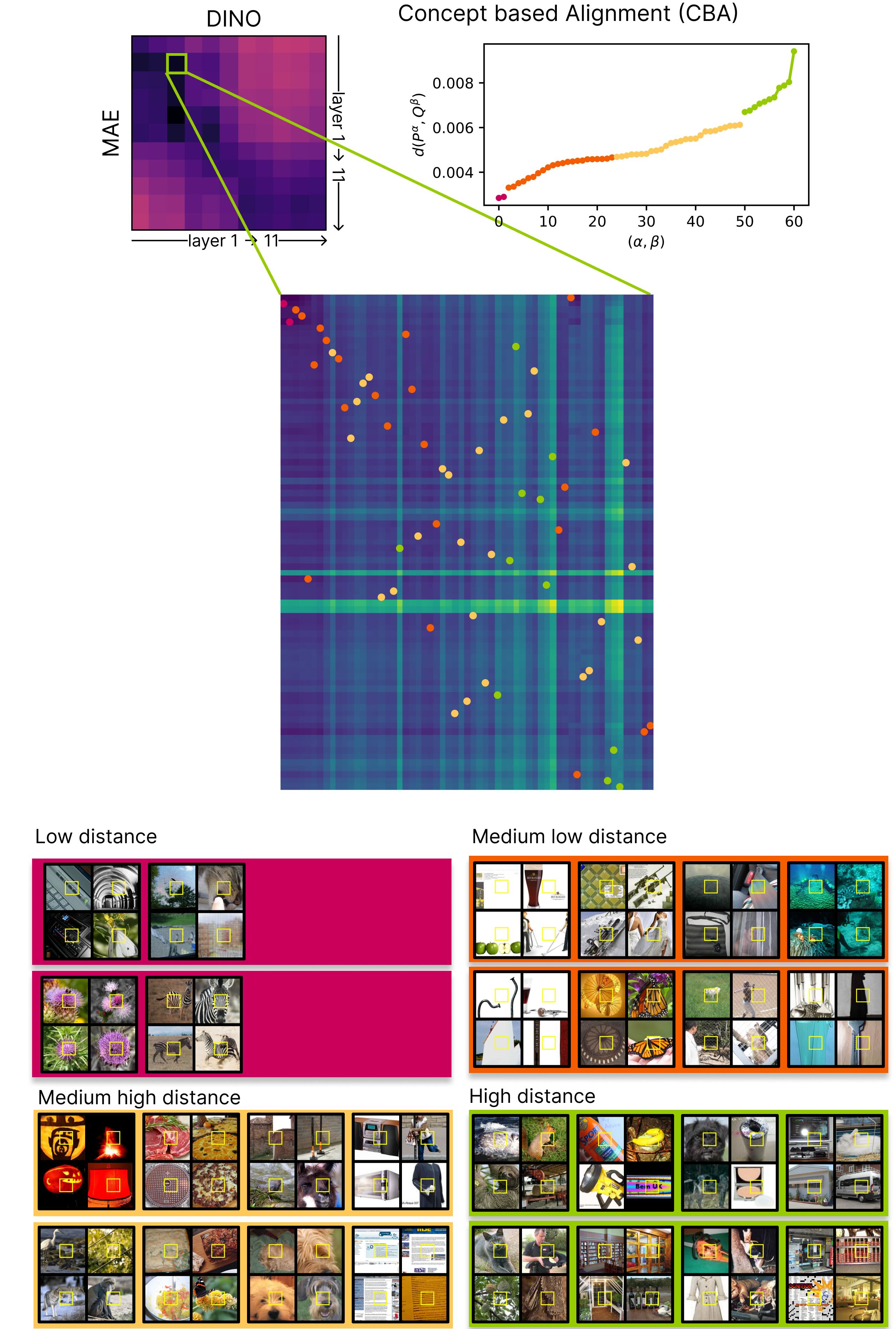}
    \caption{ CBA of SEQ concepts across layers of MAE and DINO (darker pixels correspond to higher alignment). We zoom in into the concept-wise distances $d_{cross}(P^{\alpha},P^{\beta} )$  between the representation of layer 2 in MAE and layer 3 in DINO. We give examples of universal concepts with low distance and unique ones with high distance from matches of pairs that minimize the total distance.}
    \label{fig:inter-model_SEQ1}
\end{figure}

\begin{figure}[h]
    \centering
    \includegraphics[width=\linewidth]{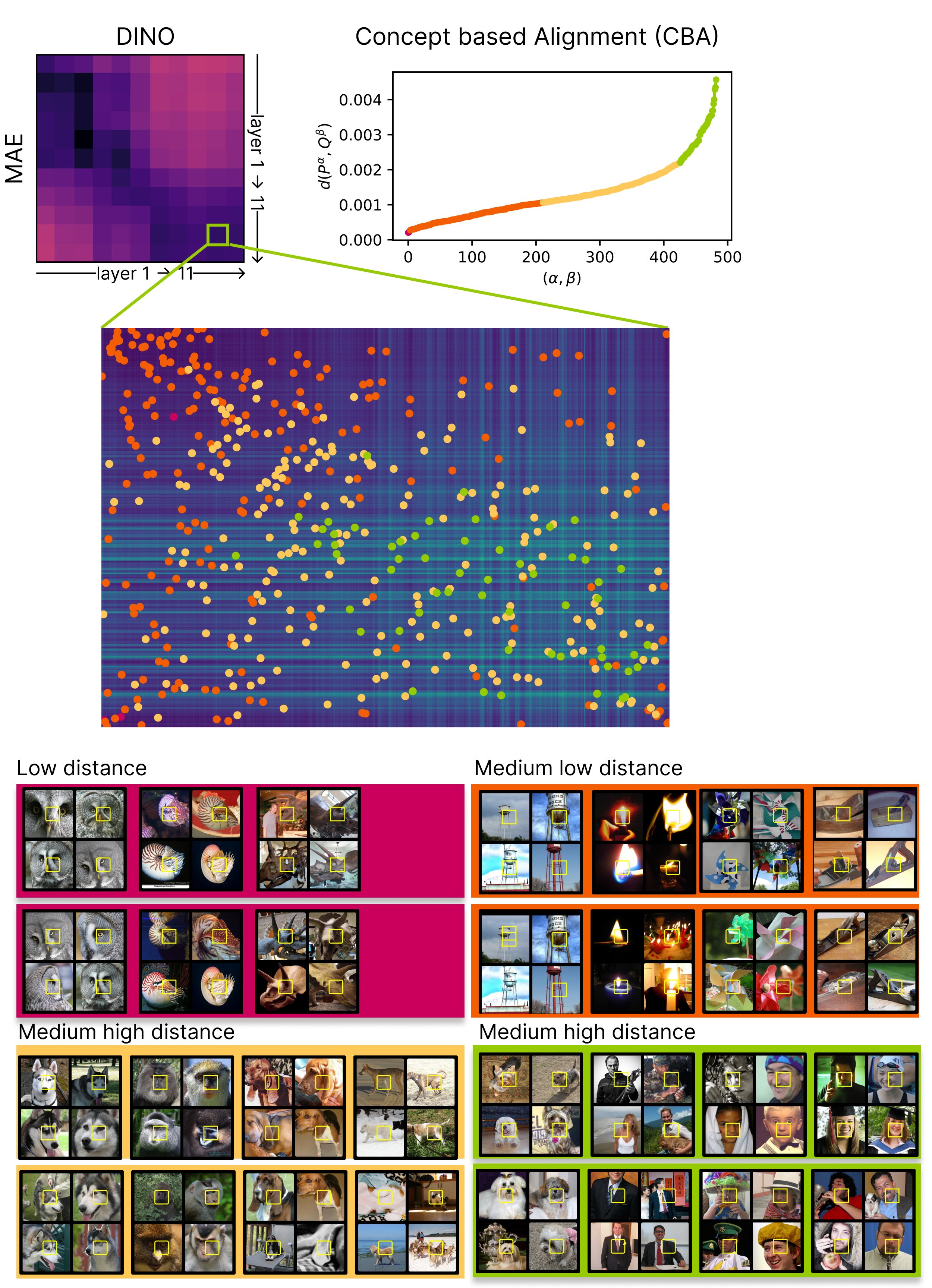}
    \caption{  CBA of SEQ concepts across layers of MAE and DINO (darker pixels correspond to higher alignment). We zoom in into the concept-wise distances $d_{cross}(P^{\alpha},P^{\beta} )$  between the representation of layer ten in MAE and DINO. We give examples of universal concepts with low distance and unique ones with high distance from matches of pairs that minimize the total distance.}
    \label{fig:inter-model_SEQ2}
\end{figure}

\paragraph{Inter-model} In the main paper, we show fine-grained inter-model concept distances between DINO and MAE at layer six in the center of the both models. Here, we add fine-grained concept distance anaylsis in the first and last part of the models in \cref{fig:inter-model_SEQ1} and \cref{fig:inter-model_SEQ2}. 
We show the full pairwise distance matrix as well as how distances between matched pairs are distributed. We partition the pairs into four regimes of distances with low, medium-low, medium-high, and high distance. 
The concept pairs between MAE layer 3 and DINO layer 2 seem to correspond mostly to edge detectors or abstract patterns. Among the low and medium-low distance pairs are grid vs. stripes (second low-distance concept pair) and diagonal edge detectors (seond medium-low distance concept pair). The common nature of the medium-high to high-distance pairs is hard to interpret but pairs include warm vs. bright light (first medium-high distance concept pair), and blurriness (third high-distance concept pair). The limitation of visualizing low-level concepts through ImageNet-1k images, as described in the main paper, becomes apparent here.
In contrast, the matched concept pairs between MAE and DINO layer 10 are easier to interpret - e.g. owl face or flame.

\begin{table*}[h]
\centering
\tiny
    \begin{tabular}{p{1.2cm} p{15cm}}
\toprule
 Category & ImageNet-1k class \\
\midrule
amphibian & European fire salamander axolotl bullfrog common newt eft
 spotted salamander tailed frog tree frog \\
artifacts & Afghan hound Band Aid Dutch oven Petri dish abacus ashcan
 backpack ballpoint bannister barrel bath towel bathtub
 beacon beaker beer bottle beer glass bell cote binder
 birdhouse book jacket bottlecap brass breakwater breastplate
 broom bucket cannon carousel carton cassette chain mail
 chainlink fence chiffonier cleaver cliff dwelling cloak clog
 cocktail shaker coffee mug comic book cowboy boot crate crib
 crutch cuirass cup diaper dishwasher dock envelope
 espresso maker face powder fig fire screen flagpole fountain
 fountain pen gasmask goblet grasshopper grille grocery store
 guillotine hair spray hand blower holster honeycomb iron
 "jack-o-lantern" joystick ladle lampshade lens cap library
 lipstick lotion mailbag mailbox manhole cover mask
 matchstick maze measuring cup megalith menu microwave
 minibus mixing bowl mobile home mortar mortarboard
 mosquito net mountain tent muzzle necklace obelisk packet
 paddle patio pedestal pencil box pencil sharpener perfume
 pickelhaube picket fence pier piggy bank pill bottle pillow
 pitcher plastic bag plate rack pole pop bottle pot
 prayer rug purse quill quilt racket radio rain barrel
 refrigerator rotisserie rubber eraser running shoe safe
 saltshaker scabbard school bus schooner scoreboard shield
 shoji shopping basket shower curtain ski mask sleeping bag
 sliding door soap dispenser soup bowl space bar spotlight
 steel arch bridge stone wall stove street sign stretcher
 sunscreen suspension bridge swab swing teddy television
 thatch theater curtain thimble tile roof totem pole
 traffic light tray triumphal arch trolleybus tub turnstile
 umbrella vacuum vase viaduct waffle iron washbasin washer
 water bottle water jug water tower web site whiskey jug
 window screen window shade wine bottle worm fence wreck yurt \\
bird & African grey American coot American egret European gallinule
 albatross bald eagle bee eater bittern black grouse
 black stork black swan brambling bulbul bustard chickadee
 cock coucal crane dowitcher drake flamingo goldfinch
 goose great grey owl hen hornbill house finch hummingbird
 indigo bunting jacamar jay junco king penguin kite limpkin
 little blue heron lorikeet macaw magpie ostrich oystercatcher
 partridge pelican prairie chicken ptarmigan quail
 red-backed sandpiper red-breasted merganser redshank robin
 ruddy turnstone ruffed grouse spoonbill sulphur-crested cockatoo
 toucan vulture water ouzel white stork \\
building & apiary barn boathouse castle church cinema greenhouse
 home theater monastery mosque palace planetarium prison
 restaurant stage stupa vault \\
canine & African hunting dog Airedale American Staffordshire terrier
 Appenzeller Arctic fox Australian terrier Bedlington terrier
 Bernese mountain dog Blenheim spaniel Border collie
 Border terrier Boston bull Bouvier des Flandres Brabancon griffon
 Brittany spaniel Cardigan Chesapeake Bay retriever Chihuahua
 Dandie Dinmont Doberman English foxhound English setter
 English springer EntleBucher Eskimo dog French bulldog
 German short-haired pointer Gordon setter Great Dane
 Great Pyrenees Greater Swiss Mountain dog Ibizan hound
 Irish setter Irish terrier Irish water spaniel Irish wolfhound
 Italian greyhound Japanese spaniel Kerry blue terrier
 Labrador retriever Lakeland terrier Leonberg Lhasa Maltese dog
 Mexican hairless Newfoundland Norfolk terrier Norwegian elkhound
 Norwich terrier Pekinese Pembroke Pomeranian
 Rhodesian ridgeback Rottweiler Saint Bernard Saluki Samoyed
 Scotch terrier Scottish deerhound Sealyham terrier
 Shetland sheepdog Shih-Tzu Siberian husky
 Staffordshire bullterrier Sussex spaniel Tibetan mastiff
 Tibetan terrier Walker hound Weimaraner Welsh springer spaniel
 West Highland white terrier Yorkshire terrier affenpinscher
 basenji basset beagle black-and-tan coonhound bloodhound
 bluetick borzoi briard bull mastiff cairn chow clumber
 cocker spaniel collie coyote curly-coated retriever dalmatian
 dhole dingo flat-coated retriever giant schnauzer
 golden retriever grey fox groenendael hyena keeshond kelpie
 kit fox komondor kuvasz malamute malinois miniature pinscher
 miniature poodle miniature schnauzer otterhound papillon pug
 red fox red wolf redbone schipperke silky terrier
 soft-coated wheaten terrier standard poodle standard schnauzer
 timber wolf toy poodle toy terrier vizsla whippet white wolf
 wire-haired fox terrier \\
clothing & Christmas stocking Loafer Old English sheepdog Windsor tie
 abaya academic gown apron bathing cap bearskin bib bikini
 bolo tie bonnet bow tie brassiere bulletproof vest cardigan
 chest cowboy hat crash helmet dishrag feather boa fur coat
 gown handkerchief hook hoopskirt jean jersey kimono
 knee pad lab coat maillot military uniform miniskirt mitten
 overskirt pajama paper towel poncho sandal sarong seat belt
 shower cap sock sombrero stole suit sweatshirt
 swimming trunks trench coat velvet vestment wallet wig wool \\
device & accordion acoustic guitar analog clock assault rifle banjo
 barometer bassoon binoculars bow buckle bullet train candle
 car mirror car wheel cash machine cello chime
 combination lock desktop computer digital clock digital watch
 disk brake drum drumstick electric fan electric guitar flute
 gas pump gong hair slide hammer hamper hand-held computer
 hard disc harmonica harp hatchet horn hourglass laptop
 lighter loudspeaker loupe magnetic compass maraca marimba
 maypole microphone missile monitor mouse mousetrap
 neck brace notebook oboe odometer oil filter organ
 oxygen mask paddlewheel padlock paintbrush panpipe
 parking meter pick "potters wheel" projector puck radiator
 radio telescope remote control revolver rifle safety pin sax
 scale screen sewing machine ski slide rule slot slug
 snorkel solar dish space heater spider web steel drum
 stethoscope stopwatch strainer sundial sunglass sunglasses
 switch syringe thresher toaster torch tripod trombone
 typewriter keyboard upright vending machine violin wall clock
 whistle \\
equipment & CD player Polaroid camera balance beam barbell "carpenters kit"
 cassette player cellular telephone computer keyboard croquet ball
 crossword puzzle dial telephone drilling platform dumbbell
 golf ball golfcart horizontal bar iPod jigsaw puzzle modem
 oscilloscope parachute parallel bars pay-phone photocopier
 ping-pong ball plate punching bag reel reflex camera
 soccer ball tape player \\
establishment & bakery barbershop bookshop butcher shop confectionery
 shoe shop tobacco shop toyshop \\
feline & Egyptian cat Persian cat Siamese cat catamount cheetah coil
 cougar leopard lion panther snow leopard tabby tiger
 tiger cat \\
fish & anemone fish barracouta coho eel electric ray gar goldfish
 great white shark hammerhead lionfish puffer rock beauty
 stingray sturgeon tench tiger shark \\
food & French loaf bagel burrito carbonara cheeseburger
 chocolate sauce consomme cucumber dough eggnog espresso
 guacamole hay hot pot hotdog ice cream ice lolly
 mashed potato meat loaf pizza potpie pretzel red wine
 trifle \\
fruit & Granny Smith acorn buckeye hip jackfruit lemon orange
 pineapple rapeseed strawberry \\
furniture & altar barber chair bassinet beaver bookcase china cabinet
 cradle desk dining table entertainment center file
 folding chair four-poster medicine chest milk can mink otter
 park bench pool table rocking chair studio couch table lamp
 throne toilet seat wardrobe \\
geological formation & alp bubble cliff coral reef dome geyser lakeside
 promontory sandbar seashore valley volcano \\
insect & ant bee cabbage butterfly cicada cricket damselfly
 dragonfly dung beetle fly ground beetle lacewing ladybug
 leaf beetle leafhopper long-horned beetle lycaenid mantis
 monarch peacock rhinoceros beetle ringlet sulphur butterfly
 tiger beetle walking stick weevil \\
mammal & African elephant American black bear Angora Arabian camel
 Indian elephant Madagascar cat Sus scrofa armadillo baboon
 bighorn bison black-footed ferret brown bear capuchin
 chimpanzee colobus dugong echidna fitch fox squirrel
 gazelle gibbon gorilla grey whale guenon guinea pig hamster
 hare hartebeest hippopotamus ibex ice bear impala indri
 killer whale koala langur lesser panda llama macaque
 marmoset marmot meerkat mongoose orangutan ox panda patas
 platypus polecat porcupine proboscis monkey ram sea lion
 siamang sloth bear spider monkey squirrel monkey
 three-toed sloth titi tusker wallaby warthog water buffalo
 wombat wood rabbit zebra \\
material & chain cornet doormat groom knot spindle toilet tissue \\
musical & grand piano \\
organism & American lobster Dungeness crab German shepherd admiral agaric
 badger ballplayer barn spider black and gold garden spider
 black widow bolete boxer brain coral centipede chiton
 cockroach conch coral fungus crawfish dam ear earthstar
 fiddler crab flatworm garden spider gyromitra harvester
 harvestman hen-of-the-woods hermit crab hog howler monkey
 isopod jellyfish king crab mushroom nematode nipple printer
 rock crab rule scorpion scuba diver sea cucumber sea slug
 sea urchin snail spiny lobster starfish stinkhorn tarantula
 tick trilobite weasel wing wolf spider \\
plant & acorn squash artichoke banana bell pepper broccoli
 butternut squash cardoon cauliflower corn custard apple daisy
 head cabbage ocarina pinwheel pomegranate sea anemone sorrel
 spaghetti squash "yellow ladys slipper" zucchini \\
reptile & African chameleon African crocodile American alligator
 American chameleon Gila monster Indian cobra Komodo dragon
 agama alligator lizard banded gecko boa constrictor box turtle
 common iguana diamondback frilled lizard grass snake
 green lizard green mamba green snake hognose snake king snake
 leatherback turtle loggerhead mud turtle night snake
 ringneck snake rock python sand viper sea snake sidewinder
 terrapin thunder snake triceratops vine snake water snake
 whiptail \\
sport & baseball basketball football helmet rugby ball tennis ball
 volleyball \\
tool & can opener chain saw corkscrew lawn mower letter opener
 lumbermill nail plane plow plunger power drill screw
 screwdriver shovel \\
utensil & Crock Pot caldron coffeepot frying pan spatula teapot wok
 wooden spoon \\
vehicle & Model T aircraft carrier airliner airship ambulance amphibian
 balloon barrow beach wagon bicycle-built-for-two bobsled cab
 canoe catamaran chambered nautilus container ship convertible
 dogsled electric locomotive fire engine fireboat forklift
 freight car garbage truck go-kart gondola half track
 horse cart jeep jinrikisha lifeboat limousine liner minivan
 moped motor scooter mountain bike moving van oxcart
 passenger car pickup pirate racer recreational vehicle
 shopping cart snowmobile snowplow space shuttle speedboat
 sports car steam locomotive streetcar submarine tank
 tow truck tractor trailer truck tricycle trimaran unicycle
 wagon warplane yawl \\
\bottomrule
\end{tabular}

    \caption{Mapping between categories from the WordNet Hierarchy and the ImageNet-1k classes used for assigning a category to the concept clusters.}
    \label{tab:wordnet_mao}
\end{table*}

\end{document}